\PassOptionsToPackage{a4paper,top=1in,bottom=1in,left=1in,right=1in}{geometry}
\documentclass[final,1p,times, 12pt]{elsarticle}

\usepackage{amssymb}
\usepackage{amsmath}
\usepackage{graphicx}
\usepackage{hyperref}
\usepackage[labelfont=bf]{caption} 
\usepackage{float}
\usepackage{bm}
\journal{Neural Networks}

\makeatletter
\def\ps@pprintTitle{%
    \let\@oddhead\@empty
    \let\@evenhead\@empty
    \def\@oddfoot{\footnotesize\hfill\today} 
    \let\@evenfoot\@oddfoot}
\makeatother

\date{\today}  

\begin{document}

\begin{frontmatter}

\title{Mitigating Spectral Bias in Neural Operators via High-Frequency Scaling for Physical Systems}

\author[a]{Siavash Khodakarami}
\author[b]{Vivek Oommen}
\author[a]{Aniruddha Bora}
\author[a,c]{George Em Karniadakis\corref{cor1}}
\cortext[cor1]{Corresponding Author: george\_karniadakis@brown.edu}

\affiliation[a]{organization={Division of Applied Mathematics, Brown University},  
                city={Providence},  
                state={RI}, 
                postcode={02912},
                country={USA}}

\affiliation[b]{organization={School of Engineering, Brown University}, 
                city={Providence}, 
                state={RI}, 
                postcode={02912},
                country={USA}}
\affiliation[c]{organization={Pacific Northwest National Laboratory},
            city={Richland},
            state={WA},
            postcode={99354},
            country={USA}}

\begin{abstract}
Neural operators have emerged as powerful surrogates for modeling complex physical problems. However, they suffer from spectral bias making them oblivious to high-frequency modes, which are present in multiscale physical systems. Therefore, they tend to produce over-smoothed solutions, which is particularly problematic in modeling turbulence and for systems with intricate patterns and sharp gradients such as multi-phase flow systems. In this work, we introduce a new approach named high-frequency scaling (HFS) to mitigate spectral bias in convolutional-based neural operators. By integrating HFS with proper variants of UNet neural operators, we demonstrate a higher prediction accuracy by mitigating spectral bias in single and two-phase flow problems. Unlike Fourier-based techniques, HFS is directly applied to the latent space, thus eliminating the computational cost associated with the Fourier transform. Additionally, we investigate alternative  spectral bias mitigation through diffusion models conditioned on neural operators. While the diffusion model integrated with the standard neural operator may still suffer from significant errors, these errors are substantially reduced when the diffusion model is integrated with a HFS-enhanced neural operator.
\end{abstract}


\end{frontmatter}



\section{Introduction}
\label{section1_1}

Design and control problems in engineering often require repeated simulation of the underlying physical system, necessitating the solution of governing partial differential equations (PDEs) multiple times.
For a wide range of applications from fluid dynamics to material science, classical discretization-based direct numerical simulation (DNS) \citep{godunov1959finite, eymard2000finite, karniadakis2005spectral, hughes2012finite} has been the cornerstone of scientific computing.
While the methods for DNS have matured significantly over the past several decades, their computational cost becomes prohibitive when performing repeated simulations over varying parametric conditions or configurations.
This challenge has fueled a growing interest in developing computationally efficient surrogate models capable of approximating these simulations at only a fraction of the cost.\\

In particular, the classical DNS can estimate the solution for a given set of conditions. 
If one of these conditions is modified, the solver has to be re-run, further aggravating the computational cost. 
To mitigate this issue, neural operators were developed to handle a plurality of conditions and parametric settings \citep{Lu2021, li2020fourier, cao2024laplace, tripura2023wavelet, ovadia2023real, li2022transformer, sharma2024graph}. 
Neural operators, which are based on the universal operator approximation theorem \citep{approx_the}, are trained to learn the mapping between infinite-dimensional functional spaces. 
Although it is expensive to train such surrogates offline, a trained neural operator can efficiently estimate solutions of unseen conditions almost instantaneously during inference. 
Many studies have used neural operators as surrogates to learn physical problems in space and time. 
Various physical problems such as vortex-induced vibration \citep{wan2025deepvivonet}, crack nucleation and propagation \citep{kiyani2024predicting}, Riemann problems \citep{peyvan2024riemannonets}, turbulence \citep{li2023long, jiang2025implicit}, plasma modeling \citep{plasma}, and many more have been solved, at least under limited conditions, by neural operators. 
Furthermore, other studies \citep{montes2021accelerating, oommen2022learning, oommen2024rethinking} attempted to learn the temporal evolution of two-phase microstructures in diffusion-driven processes such as spinodal decomposition and dendritic growth. 
However, very few studies have investigated the application of neural operators for buoyancy-dominated or advection-dominated two-phase flow problems, such as those encountered in boiling and condensation \citep{khodakarami2023intelligent}.\\ 

\subsection{Neural operators and applications in two-phase flow modeling}
\label{subsection1}
Modeling and predicting two-phase flow during boiling is one of the most challenging problems in computational fluid dynamics. 
These phenomena involve complex interface dynamics and phase transitions, resulting in high-frequency spatio-temporal variations that are both challenging and computationally expensive to capture. 
Analyzing the solutions of such a system reveals a slowly decaying energy spectrum, where even the high wavenumbers carry a non-trivial amount of energy that cannot be neglected.  
Effective modeling of a two-phase flow system requires the neural operators to accurately predict spatio-temporal evolution of both low and high wavenumber modes. 
Unfortunately, neural networks and neural operators suffer from spectral bias \citep{spectral_bias,spectral2, xu2025understanding}, which makes them oblivious to high wavenumber modes.
Consequently, the neural operators can only offer an over-smoothed prediction that fails to capture the intricate features near the interfaces where the sharp gradients are commonly observed.\\

Previous studies in boiling modeling with neural operators also confirm the spectral bias problem. 
\cite{bubble_DeepONet} used DeepONet \citep{Lu2021} to solve for the transient solution of a single bubble growth. Their findings demonstrate that DeepONet can effectively capture the mean component of the solution in the microscale regime, but it fails to accurately predict the stochastic fluctuations described by high-frequency components of the solution. 
A study by Jain et al. \cite{mp_flow_simple} on the prediction of multiphase flow through porous media with UNet \citep{ronneberger2015u} also showed that larger errors occurred near the interfaces.
The Fourier neural operator (FNO) \citep{li2020fourier} also suffers from spectral bias \citep{FNO_spectral}. 
The common practice of truncating high-frequency modes in FNOs leads to the loss of rich information, hindering the accurate modeling of chaotic systems in multi-phase heat transfer and turbulence. 
However, without truncation, training FNOs becomes unstable \citep{ovadia2023real}.\\

A recent study by Hassan et al.\cite{bubbleML} collected a valuable boiling dataset based on Flash-X simulations \citep{FlashX} and developed neural operators based on different structures such as UNet, FNO, and group equivariant FNO (GFNO) for prediction in boiling problems. As shown in the results of our work, the previously best neural operator still struggles to capture high-frequency modes, which are prominently observed within the bubbles, along the interfaces, and in condensation traces in subcooled pool boiling. 
These over-smoothened solutions highlight the need for further advancements to mitigate spectral bias in modeling phase-change and multi-phase flow phenomena. Similarly, spectral bias of neural operators cannot be overlooked when modeling other chaotic systems like turbulence \citep{turbulence_spectral}, where small-scale, low-energy features play a crucial role.

\subsection{Spectral bias mitigation strategies}
\label{subsection1_2}
Previous studies have proposed various methods to mitigate spectral bias and over-smoothing in deep neural networks (DNNs). Cai et al. \cite{cai2019multi} proposed a multi-scale DNN (MscaleDNN) to enhance approximations over a wide range of frequencies for the solution of PDEs. Tancik et al. \cite{tancik2020fourier} proposed Fourier feature mapping for coordinate-based multilayer perceptron (MLP) to tackle spectral bias in image regression tasks in low dimensional domains. Wang et al. \cite{wang2021eigenvector} used Fourier feature mapping along with Physics-informed Neural Networks (PINNs) \citep{raissi2019physics, toscano2024pinns} to enhance the multi-scale PDE solutions by mitigating the spectral bias compared to vanilla PINN. A better optimization of activation functions have been also shown to slightly reduce spectral bias of DNNs and PINNs \citep{liang2021reproducing,jagtap2020adaptive}. Phase shift DNN is another method converting high-frequency component of the data into low frequency spectrum, which can be learned and represented by a DNN. Subsequently, the learned representation is converted into the original 
high-frequency. However, phase shift DNN suffers from the curse of dimensionality \citep{cai2020phase}. \\

Efforts have also been made to mitigate the spectral bias encountered by neural operators trained to learn spatiotemporal systems.
Lippe at al. \cite{lippe2023pde} developed PDE-Refiner, which iteratively adds noise to perturb different scales of the system and trains the neural operator to correct these corrupted states. 
Zhang et al. \cite{zhang2024blending} developed Hybrid Iterative Numerical Transferable Solver (HINTS) to exploit the spectral bias in
solving large linear systems by blending neural operators and relaxation methods. 
Generative Artificial Intelligence (GenAI)--based algorithms are also emerging as effective methods to overcome the spectral bias barrier. 
Wu et al. \cite{wu2024high} accurately reconstructed the small-scale structures accompanying turbulent boundary layers in wall turbulence using Super Resolution Generative Adversarial Networks (SRGANs). 
Wang et al. \cite{wang2022deep} developed a framework based on GANs to reconstruct high spatiotemporal resolution supersonic flow states from sparse measurements. 
Molinaro et al. \cite{molinaro2024generative} developed GenCFD using score-based diffusion models to learn three-dimensional turbulence in compressible and incompressible flows. 
Lockwood et al. \cite{lockwood2024generative} used denoising diffusion probabilistic models to refine the estimates of tropical cyclone wind intensities. 
Oommen et al. \cite{oommen2024integrating} addressed the spectral limitations of neural operators in modeling a series of turbulent systems by training a conditional score-based diffusion model conditioned on the neural operator as prior.\\ 

In this work, we first propose the use of UNet with residual blocks (ResUNet) to achieve more accurate two-phase flow predictions compared to the state-of-the-art neural operators. Subsequently, we present a new method named high-frequency scaling (HFS) to mitigate spectral bias in two-phase flow predictions. Our approach demonstrates higher accuracy and better alignment of energy spectra, with negligible additional memory requirements and only a small computational overhead on the neural operator. We applied HFS to different variants of ResUNet. Finally, we explore the dependency of diffusion models on the prior accuracies when integrated with neural operators. Specifically, we show that the integration of the diffusion model with neural operators equipped with HFS results in further mitigation of spectral bias without compromising prediction accuracy.
We demonstrate the effectiveness of our methodology for both two-phase and single-phase flows.\\

The manuscript is organized as follows. 
We start with providing an in-depth description about neural operators, HFS, and diffusion models in Section \ref{sec:methods}. 
We present the results of our investigations in Section \ref{section3}, followed by discussion and summary in Sections \ref{Discussion} and \ref{Summary}, respectively. In the Apendix, we include more technical details and additional results.


\section{Methods}
\label{sec:methods}
\subsection{Neural Operators}
\label{subsection2_1}



The mathematical operator $\mathcal{N}$ that governs the temporal evolution of a time-dependent system can be expressed as,
\begin{equation}
    \bm{u}(\bm{x}, t + \Delta t) \approx \mathcal{N}(\bm{u}(\bm{x}, t)) (\Delta t),
\end{equation}
where $\bm{u}$ is the representative state variable(s) of interest. The objective here is to train a neural operator $\mathcal{F}_{\theta}$ to learn the true underlying operator ($\mathcal{N}$) by, typically, minimizing the mean of an error norm such as $||\bm{u}(\bm{x}, t+\Delta t) - \mathcal{F}_{\theta}(u(\bm{x}, t))(\Delta t)||_2$.\\

In this work, we focus on resolving solutions in pool boiling problems and single-phase turbulent flows. We start our analysis with pool boiling problems. Then, we investigate the application of our method on single-phase turbulent flows. There have been several efforts to use neural operators to learn temperature and flow dynamics in two-phase flow problems. Here, we demonstrate the advantage of using the ResUNet structure compared to previously developed neural operators such as UNet and FNO for two-phase flow problems with high-frequency features \citep{hassan2023bubbleml}. The models are trained to predict future temperatures based on temperature history and velocity information. The problem configuration is shown in Equation 2, where $\bm{x}$ is the spatial mesh, $T$ is the temperature, $V$ is the velocity, $k$ specifies the prediction time interval length, and \(\mathcal{F}_{\mathrm{\theta}}\) is the trained neural operator.

\begin{equation}
    T(\bm{x}, t:t + k\Delta t) = \mathcal{F}_{\mathrm{\theta}}(T(\bm{x}, t - k\Delta t:t),V(\bm{x}, t-k\Delta t:t + k\Delta t))
\end{equation}

UNet with residual blocks (ResUNet) was first introduced for a semantic segmentation task by imposing skip connections between convolutional layers in a UNet-like structure \citep{diakogiannis2020resunet}. We use the same idea to add skip connections in the form of residual blocks to both the encoder and decoder side of the UNet. The residual blocks have been shown to mitigate vanishing gradient problems by offering a smoother optimization landscape \citep{li2018visualizing}. We also demonstrate that they help with the better flow of information in the network for complex datasets such as two-phase flows, which results in a better capture of localized features, possibly reducing the spectral bias towards low-frequency components. We also introduced several  modifications, such as the GELU activation function and group normalization, that demonstrated superior prediction accuracy. We used the mean squared error (MSE) loss function in all prediction time steps (Equation 3) as the objective criterion to train the model, i.e., 
\begin{equation}
    L(\theta) = \frac{1}{N_uk}\sum_{i=1}^{N_u}\sum_{j=1}^{k}\|T^i(\bm{x}, t+j\Delta t) - \mathcal{F}_{\theta}(T^i(\bm{x}, t))(j\Delta t) \|_2^2
\end{equation}\\
We used the Lion optimizer \citep{chen2024symbolic} to perform the optimization as we observed superior performance with this optimizer compared to the conventional Adam optimizer \citep{kingma2014adam}. More details about the ResUNet structure, the training hyperparameters, and comparison with UNet predictions are included in Appendix A.

We evaluated our baseline neural operator on both saturated and subcooled pool boiling datasets from the BubbleML data repository, which is generated through Flash-X simulations \citep{dubey2022flash} and were collected in a previous study \citep{hassan2023bubbleml}. It should be noted that predictions in subcooled boiling is more difficult due to the vortices generated by condensation trails. Therefore, the errors are higher in subcooled boiling predictions, and the results look more over-smoothed compared to saturated boiling prediction results. A visualization of the subcooled boiling prediction results is shown in \ref{appA}. A comprehensive comparison of our baseline model with the previous best baseline model developed by \cite{hassan2023bubbleml} is included in Table 1 and Table 2 for saturated and subcooled pool boiling dataset, respectively. The ResUNet improves the resolution of high-frequency features, resulting in higher prediction accuracy.
We note that given the possible differences in the testing dataset, the one-to-one comparison with the previously reported numbers may not be fair. Therefore, we trained and tested the previously reported best model (e.g., UNet) with our dataset configuration, which consists of a larger test dataset and smaller training dataset compared to the previous work.\\

We evaluated our model using six different field metrics relevant to two-phase problems. These metrics include relative error (Rel. Error), root mean square error (RMSE), boundary RMSE (BRMSE) showing the error on the boundaries, bubble RMSE showing the error in the bubble areas and at the interfaces, mean maximum error (\( \text{Max}_{\text{mean}} \)) showing the mean of the maximum error for each prediction, and overall maximum error (\( \text{Max}_{\text{max}} \)) showing the maximum error over the test dataset. We also evaluated the predictions in three different frequency bands using spectral errors at low frequency (\(F_\text{low}\)), medium frequency (\(F_\text{mid}\)), and high frequency (\(F_\text{high}\)). Exact definitions of BRMSE and bubble RMSE, as well as spectral errors are described in Appendix B. All the metrics are computed on the normalized dataset ($T^i(\bm{x},t+j\Delta t) \in [-1,1] \text{ } \forall \text{ } \{i,j\}$). 
For all the results, the state of the temperature at five future time-steps are predicted based on five time-step previous temperature history and the velocity information in two spatial dimensions.

\begin{table}[H]
    \caption{\textbf{Saturated pool boiling temperature prediction errors.} The training dataset consists of simulations from 11 different wall temperatures. The test dataset consists of simulations with two other wall temperatures (70\textdegree C, and 95\textdegree C) not seen during training.}
    \label{Table 1}
    \centering
    \begin{tabular}{c|c|c}
    \hline
        & \textbf{UNet} & \textbf{ResUNet} \\
    \hline
    \textbf{Rel. Error} &0.0191 &0.0149 \\ \hline
    \textbf{RMSE} &0.0189 &0.0148 \\ \hline
    \textbf{BRMSE} &0.0582 &0.0364 \\ \hline
    \textbf{Bubble RMSE} &0.116 &0.0726\\ \hline
    \textbf{Max\textsubscript{mean}} &0.705 &0.553\\ \hline
    \textbf{Max\textsubscript{max}} &1.204 &1.154\\ \hline
    \textbf{\textit{F}\textsubscript{low}} &0.105 &0.0745\\ \hline
    \textbf{\textit{F}\textsubscript{mid}} &0.113 &0.0919\\ \hline
    \textbf{\textit{F}\textsubscript{high}} &0.0238 &0.0182\\ \hline
    \textbf{Parameters [Millions]} &7.8& 3.5\\ \hline

    \end{tabular}
\end{table}

\begin{table}[H]
    \caption{\textbf{Subcooled pool boiling temperature prediction errors.} The training dataset consists of simulations from eight different wall temperatures. The test dataset consists of simulations with two other wall temperatures (95\textdegree C, and 98\textdegree C) not seen during training.}
    \label{Table 2}
    \centering
    \begin{tabular}{c|c|c}
    \hline  
        & \textbf{UNet} & \textbf{ResUNet} \\
    \hline
    \textbf{Rel. Error} &0.0516 &0.0295 \\ \hline
    \textbf{RMSE} &0.0501 &0.0288 \\ \hline
    \textbf{BRMSE} &0.139 &0.0646 \\ \hline
    \textbf{Bubble RMSE} &0.269 &0.127\\ \hline
    \textbf{Max\textsubscript{mean}} &1.141 &0.837 \\ \hline
    \textbf{Max\textsubscript{max}} &2.279 &1.433 \\ \hline
    \textbf{\textit{F}\textsubscript{low}} &0.346 &0.157 \\ \hline
    \textbf{\textit{F}\textsubscript{mid}} &0.367 &0.197 \\ \hline
    \textbf{\textit{F}\textsubscript{high}} &0.0583 &0.0370 \\ \hline
    \textbf{Parameters [Millions]} &7.8 &3.5\\ \hline

    \end{tabular}
\end{table}

The results in Tables 1 and 2 demonstrate that all the metrics are improved by simply introducing residual blocks in the network, a better optimizer, and a better normalization. For example, there is approximately 21\% and 42\% reduction of RMSE in saturated and subcooled boiling, respectively. Interestingly, the ResUNet achieves better accuracies with less than half of the number of parameters in UNet. Most of the prediction errors occur within the bubble areas and at the condensation trails. This is due to the larger gradients in the bubble areas and around condensation trails resulting into more complex patterns that are more challenging to capture with the neural operator. This is expected as the neural operators are known to have spectral bias to low-frequency modes. The high-frequency content typically exists in regions with significant gradients such as interfaces and condensation trails. In subcooled pool boiling, departing bubbles may condense after departure, creating vortices that gradually dissipate over time. These vortices form complex structures containing higher energy at high frequencies. As a result, subcooled boiling presents greater prediction challenges compared to saturated boiling. For instance, prediction spectral errors (\(F_{\text{low}}\), \(F_{\text{mid}}\), \(F_{\text{high}}\)) are approximately two times higher in subcooled boiling, highlighting the increased complexity with the high-frequency content. \\

While the residual blocks improve the neural operator's ability to reduce field errors (e.g., RMSE) and over-smoothing of certain high-frequency contents, the results still suffer from significant over-smoothing (see \ref{appA}). Previous studies have also shown the over-smoothing issue of convolutional based neural operators for image generation tasks and scientific computing \citep{wei2023super,wang2020towards}. Other studies demonstrated the frequency selectiveness of convolutional neural network (CNN) architechtures resulting to different learning rates for low and 
high-frequency components \citep{chakrabarty2019spectral,saxe2011random}. Wang et al. \cite{wang2022anti} demonstrated the spectral bias in vision transformers (ViT) through Fourier analysis. They showed that the problem arises by self-attention layers that act as low-pass filter and continuously reduce high-frequency information with the network depth. A feature scaling technique was proposed to decompose the attention output signal into direct and high-frequency components and scale them separately to adjust the proportion of different frequencies of the signal. We draw inspiration from this technique and propose a similar approach to separately scale low frequency and high-frequency components of the features in the latent space of the neural operator to mitigate spectral bias.

\subsection{High-frequency scaling}
\label{subsection2_2}
As discussed in Section \ref{subsection2_1}, neural operators suffer from spectral bias. While residual blocks offer improvements up to some extent, they cannot effectively mitigate the spectral bias inherent in the neural operators. Hence, we propose the high-frequency scaling (HFS) approach to be applied to the output of convolutional layers. The latent feature map of each convolutional layer is first divided into non-overlapping patches, similar to the first step in vision transformers. This will break down the spatial dimensions into smaller regions, which empirically will allow for better localized processing. We consider the mean of the patches as the direct component (DC) of these signals. Then, the high-frequency component (HFC) for each patch can be defined as the difference of each patch with the DC. It should be noted that here the DC is calculated across the patches and not individually for each patch. Then, we introduce two parameter groups of $\lambda_\text{DC}$ and $\lambda_\text{HFC}$ to separately scale the DC and HFC for each patch. We then re-assemble the patches to the original latent feature size before the next operation. 

A more rigorous description of the method is as follows: Let $X \in \mathbb{R}^{H \times W \times C}$ be the output feature map of a convolutional layer, where \(H\), \(W\), and \(C\) are the height, width, and number of channels, respectively. We divide \(X\) into \(N\) non-overlapping patches of size \(p \times p\) denoted as \(X^{(i)} \in \mathbb{R}^{p \times p \times C}\), where \(i \in [0,N]\). The DC is defined as the mean patch across all \(N\) patches as shown in Equation (4). The HFC calculation for each patch and the scaling step are shown in Equations (5-6):

\begin{equation}
    DC(X) = \frac{1}{N} \sum_{i=1}^{N} X^{(i)},
\end{equation}

\begin{equation}
    HFC(X^{(i)}) = X^{(i)} - DC(X),
\end{equation}

\begin{equation}
    \hat{X}^{(i)} = X^{(i)} + \lambda_{DC} \odot DC(X) + \lambda_{HFC} \odot HFC(X^{(i)}).
\end{equation}

The scaled feature map can be then reconstructed by re-assembling the \(\hat{X}^{(i)}\)s.

\begin{figure}[H]
    \centering
    \includegraphics[width=1\textwidth]{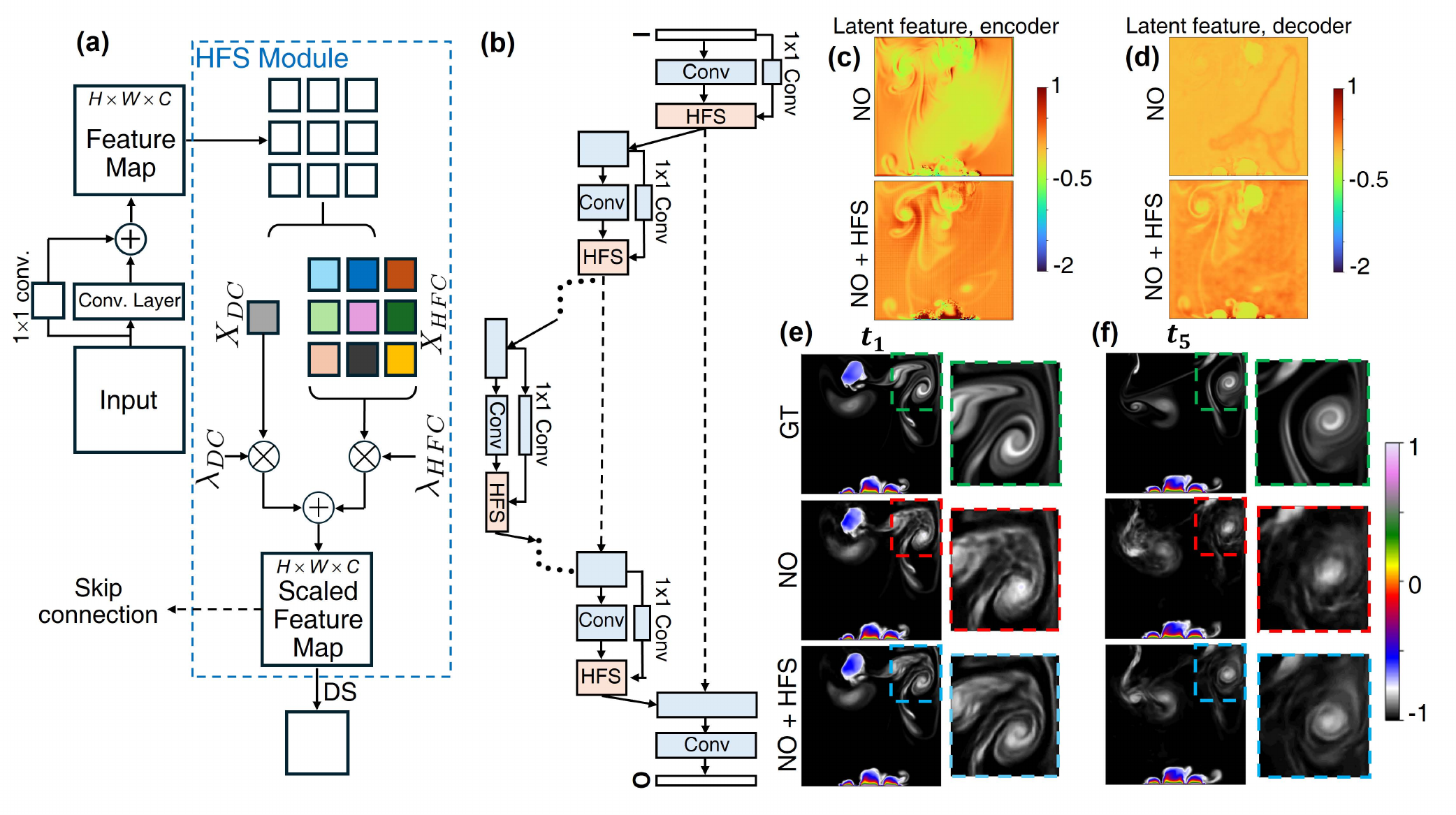}
    \caption{\textbf{Structure of the HFS-enhanced NO.} (a) Schematic of the HFS module (right) integrated with the residual block (left). (b) Structure of the ResUNet with the HFS modules (blocks in front of conv layers). (c) An example of a learned latent space feature from the first layer of the encoder trained with and without HFS. The most similar feature maps of the models in the first encoder level are shown. (d) An example of a learned latent space feature from the last layer of the decoder trained with and without HFS. The most similar feature maps of the two models at the last decoder level are shown. (e-f) Examples of temperature prediction with NO and HFS-enhanced NO at two different time-steps. A region with high-frequency features (top right corner) is zoomed in for better visualization.} 
    \label{Figure 1}
\end{figure}

The scaling parameters \(\lambda_{DC} \in \mathbb{R}^{1 \times 1 \times C}\) and \(\lambda_{HFC} \in \mathbb{R}^{1 \times 1 \times C}\) are left to be learnable parameters that are optimized using gradient descent simultaneously with the network optimization. Here, we initialized the parameters to be one and optimized them with the same learning rate used for network training. In ResUNet structure, HFS is applied to the output of both convolutional layers and the skip-connection paths with \(1 \times 1\) convolutions or identity skip-connections. In practice, HFS can be seen as a new module incorporated to each layer of the encoder and decoder, as shown in Fig. \ref{Figure 1}. Fig. \ref{Figure 1} also depicts examples of the learned feature maps for models with and without HFS. The most similar feature maps between the models from the first encoder layers and the last decoder layers are depicted. The model with HFS learns features with more pronounced high-frequency content and reduced over-smoothing, which possibly enhances the capture of high-frequency components of the solution and mitigates spectral bias of the neural operator. A summary of the improvements in prediction accuracy achieved through HFS is provided in Appendix C.

\subsection{Diffusion Model}
\label{subsection2_3}

As discussed earlier, the NO and the HFS-enhanced NO learn the solution by minimizing some variant of the Euclidean distance, such as MSE, RMSE, relative $L^2$ or relative $L^1$ norms of the errors, between the true and predicted states.
Unfortunately, such a loss function effectively prioritizes the error at those wavenumbers that bear higher energy. 
The systems considered in this study exhibit a decaying energy spectrum, implying that the lower wavenumbers carrying higher energy will be over-represented, while the higher wavenumbers that bear lower energy will be ignored due to its minimal influence on the Euclidean distance-based loss function.
The recent efforts aimed at improving the spectral bias of NO using GenAI algorithms, discussed in Section \ref{section1_1}, motivated us to explore this route.
Specifically, we investigate if diffusion models \citep{ho2020denoising} can help further refine the predictions estimated by NO and HFS-enhanced NO.\\

Diffusion models (DM) are generative frameworks capable of producing samples that align with the true underlying function distribution, $\mathcal{T}$, given a limited set of observations from $\mathcal{T}$. These models achieve sample generation by progressively refining a simple prior distribution, such as a standard normal distribution ($\Gamma_0 =\mathcal{N}(0,I)$), into the desired complex distribution ($\Gamma_N \approx \mathcal{T}$) over $N$ iterative steps. \\

The diffusion process begins with an initial sample $\mathbf{X_0}$ drawn from $\Gamma_0$ and subsequently predicts $\mathbf{X}_1$. Since $\Gamma_0 = \mathcal{N}(0,I)$, obtaining $\mathbf{X}_0$ is straightforward. The model then iteratively refines the sample, estimating $\mathbf{X}_{i+1}$ from $\mathbf{X}_i$ over $N$ steps. However, a key challenge arises on how to train the diffusion model to transition from $\Gamma_0 =\mathcal{N}(0,I)$ to $\Gamma_N \approx \mathcal{T}$ when intermediate distributions $\Gamma_i$ for $i=\{1,2,3, \dots, N-1\}$ are not explicitly available. This challenge is addressed using denoising score matching combined with Langevin dynamics \citep{song2020score}. The objective of a score-based diffusion model is to estimate the score function, which is defined as $s_{\theta_{D}}(\mathbf{X})=\nabla_{X}\log p(\mathbf{X})$, where $\theta_{D}$ represents the parameters of the diffusion model and $p$ is the probability density of $\mathbf{X}$, where $\mathbf{X}$ corresponds to continuous realizations of $\mathbf{X}_i \sim \Gamma_i$. Since the exact data distribution is unknown and may reside on a lower-dimensional manifold, the score function can become ill-defined in regions lacking data. To mitigate this issue, Gaussian noise is added to perturb the data, ensuring a well-defined score function across the entire space by smoothing the distribution. The score function provides a directional gradient toward regions of higher probability. However, a direct mechanism to sample from the learned distribution is still absent. \\

\begin{figure}[H]
    \centering
    \includegraphics[width=1.0\linewidth]{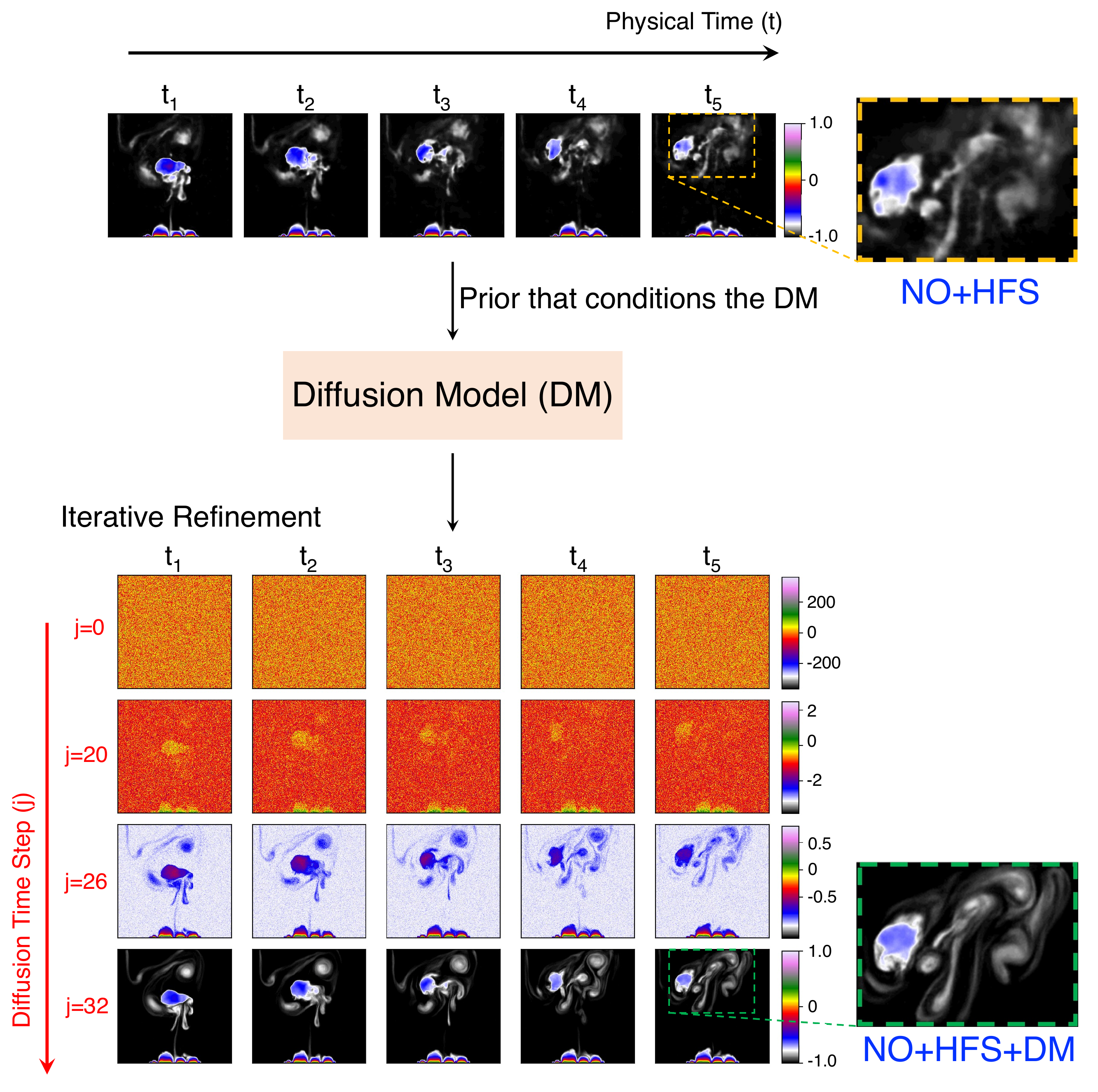}
    \caption{\textbf{Mitigating Spectral Bias with Diffusion Model.} The states estimated by the NO exhibit over-smoothing. They serve as the prior that conditions the DM, which in turn reconstructs the missing frequencies iteratively through conditional sampling. The results are based on a NO with 2 million parameters.}
    \label{fig:no_dm}
\end{figure}

This limitation is overcome using Langevin dynamics, as proposed in \citep{song2019generative}. Langevin dynamics ensures that the generated samples converge to the true underlying distribution by balancing deterministic motion, driven by the gradient of the log probability, with stochastic exploration introduced by noise. In our approach, we condition the score function on the output of the pre-trained NO, $\mathbf{\mathcal{F}_{\theta}}$, leading to the modified score function:

\begin{equation} 
\label{eq1d}  
s_{\theta_{D}}(\mathbf{X},\sigma,\mathbf{\mathcal{F}_{\theta}}) = \nabla_{X}\log p(\mathbf{X}|\mathbf{\mathcal{F}_{\theta}}),
\end{equation} 

where $\sigma$ represents the noise level. This conditioned score function guides the DM to sample from the posterior distribution of $\mathbf{X}$ given $\mathbf{\mathcal{F}_{\theta}}$, ensuring that the generated samples are consistent with both the structures imposed by $\mathbf{\mathcal{F}_{\theta}}$ and the true data distribution. The update rule for Langevin dynamics is given by:

\begin{equation} \label{eq2d}
    \mathbf{X}_{j+1} = \mathbf{X}_{j} + \frac{\varepsilon}{2}s_{\theta_{D}}(\mathbf{X}_{j},\sigma_{j},\mathbf{\mathcal{F}_{\theta}})+\sqrt{\varepsilon}z_{j},
\end{equation}

where $\varepsilon$ is the step size, $z_{j}$ is the noise component, and $\sigma_{j}$ denotes the noise scale at iteration $j$ during the sampling process. 
The iterative denoising of the noised states by a diffusion model conditioned on the outputs of a pre-trained HFS-enhanced NO is illustrated in Fig \ref{fig:no_dm}.

During training, the diffusion model learns to denoise the state of the system perturbed by a noise with zero mean and $\sigma$ standard deviation, where $\ln{\sigma} \sim \mathcal{N}(-1.2, 1.2^2)$.
When $\sigma$ is small, the score function $s_{\theta_{D}}$ increasingly focuses on reconstructing high-frequency details and vice versa.
In this manner, the diffusion model learns to perturb and reconstruct the signal at multiple scales, unlike the NO whose scale is fixed throughout its training, and thereby learns the structure of the underlying system across all the scales. 
Our implementation of the DM conditioned on the NO and HFS-enhanced NO is based on \citep{oommen2024integrating} that adopts the training, network architecture, pre-conditioning, and sampling routine proposed in ``EDM" \citep{karras2022elucidating}.

\section{Results}
\label{section3}
\subsection{HFS-enhanced NO for two-phase flow problems}
\begin{figure}[H]
    \centering
    \includegraphics[width=1\textwidth]{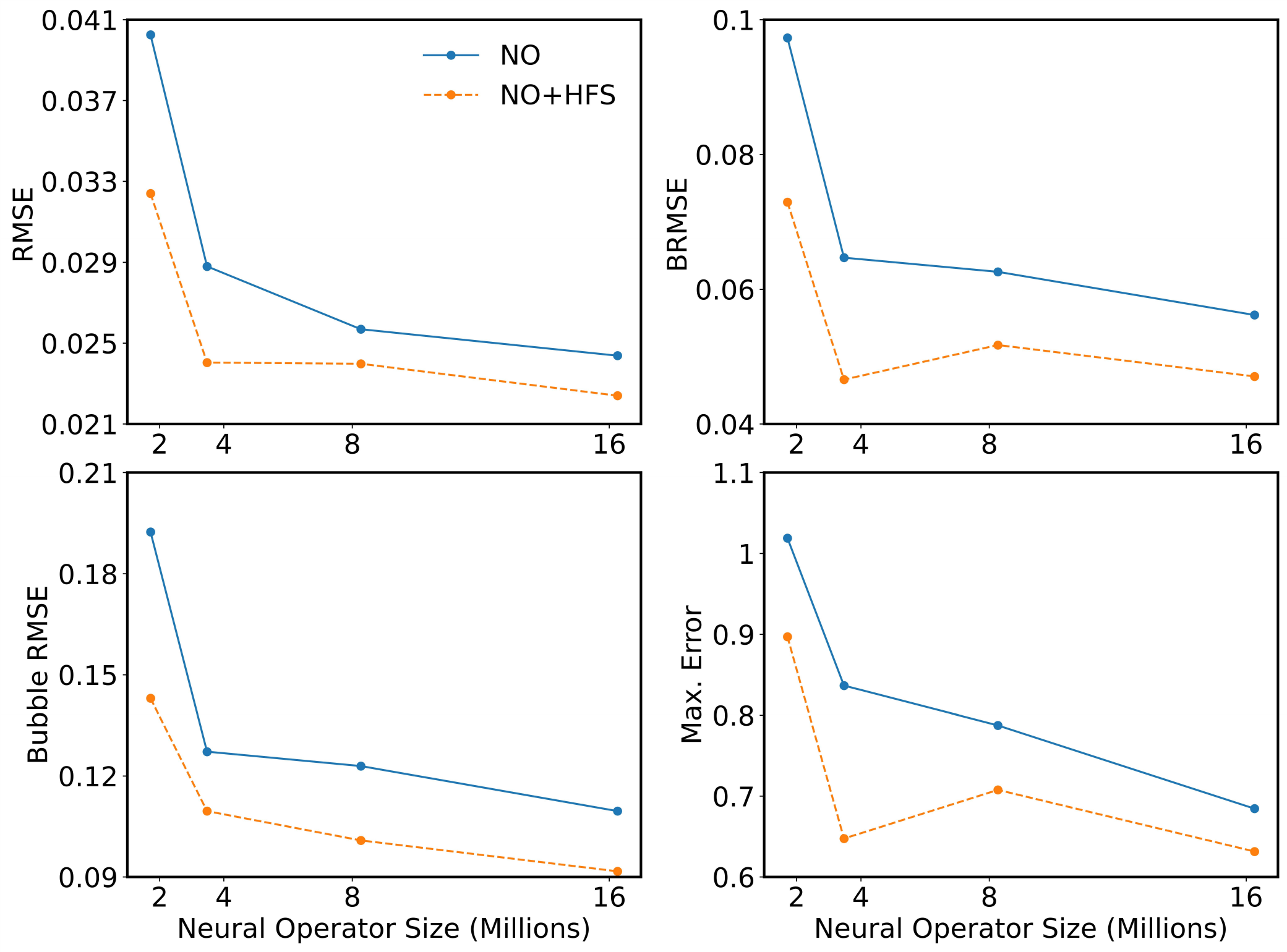}
    \caption{\textbf{Temperature prediction errors of NO and HFS-enhanced NO varying with NO size.} (a) Root mean square error (RMSE), (b) Boundary RMSE (BRMSE), (c) Bubble RMSE, (d) Mean maximum error. All the errors are calculated over the 5 time-step temperature predictions. The legends in (a) are applicable to (b - d) as well. All the results are based on test dataset in subcooled pool boiling.}
    \label{Figure 4}
\end{figure}

We first conduct several experiments with different variants of ResUNet to demonstrate the advantage of HFS in spectral bias mitigation for two-phase flow operator learning problem. Given the higher complexity of subcooled boiling data compared to saturated boiling data, we will focus on the subcooled boiling experiments. Examples showing the saturated boiling predictions are shown in Appendix D. Given the flexibility of our NO structure, we investigated different variants of ResUNet by varying the NO size by changing the number of parameters in the range of \(\sim\) 2 to 
\(\sim\) 16 million parameters.  The number of parameters was changed by simply changing the number of latent feature maps at each level of the ResUNet structure. The number of downsamplings/upsamplings was kept at five steps for all the models to achieve spatially consistent resolutions at each level across all the NOs. The subcooled pool boiling dataset consists of 10 different simulation trajectories, two of which were used for testing. Each simulation trajectory consists of 201 time-steps. However, similar to \citep{hassan2023bubbleml}, the first 30 unsteady time-steps were not included in the training and testing of the models. Fig. \ref{Figure 4} demonstrates the variation of RMSE, BRMSE, bubble RMSE, and Max\textsubscript{mean} metrics with NO size for results obtained from NO and HFS-enhanced NO. As expected, both NO and HFS-enhanced NO exhibit error-decreasing trends with the number of parameters. However, the HFS-enhanced NO always yields lower errors compared to NO in all metrics and irrespective of the NO size. The effect of HFS is more pronounced in the bubble RMSE due to larger high-frequency content at the bubble interface and within the bubbles. For example, HFS yields 8\% improvement in RMSE for the 16 million NO. This improvement is 16\% for the bubble RMSE metric. On average, HFS decreases the RMSE and bubble RMSE by 12.4\% and 18.2\%, respectively.

\subsection{Spectral analysis of HFS-enhanced NO}
\label{subsection3_2}
\begin{figure}[H]
    \centering
    \includegraphics[width=1\textwidth]{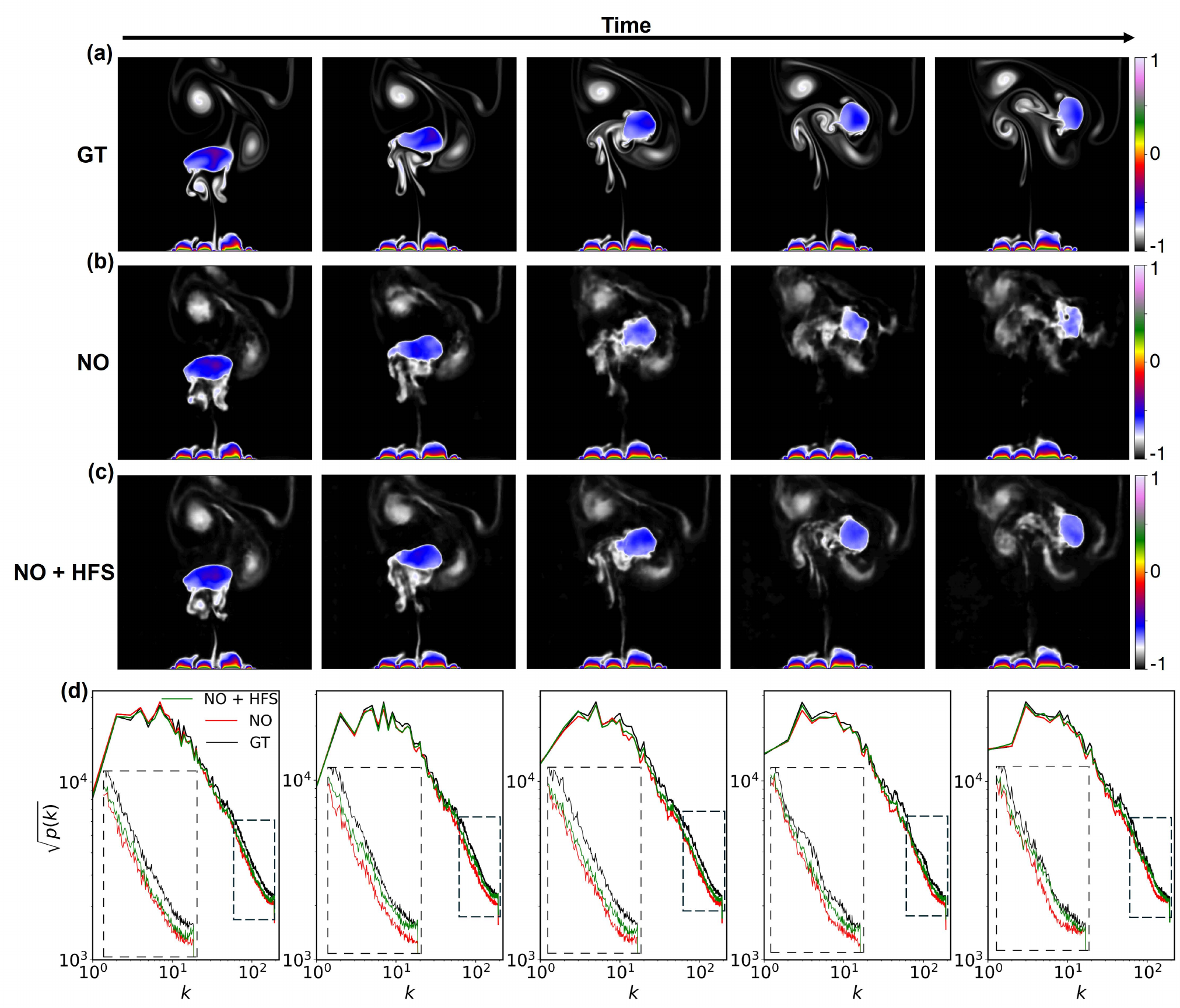}
    \caption{\textbf{Subcooled pool boiling transient temperature prediction.} (a) Ground truth (GT) temperatures for 5 consecutive time steps (from left to right) ($\Delta$\textit{t} = 8 ms). (b) NO prediction results. (c) HFS-enhanced NO prediction results. (d) The corresponding energy spectra (\textit{p}(\textit{k})) for each time step. For better visualization, the subplots in (d) show the energy spectra only for the high wavenumbers. The legends in first plot are applicable to other plots as well. All the results are based on a \(\sim\) 3.5 M parameter NO.}
    \label{Figure5}
\end{figure}

HFS reduces the over-smoothing effect,  hence, the intricate features of vortices induced by condensation trails in subcooled boiling are better resolved. Moreover, HFS results in better alignment of the energy spectra to the ground truth signal, especially at high wave numbers attributed to the high frequency features. Fig. \ref{Figure5} depicts the enhacements obtained by adding HFS modules to NO. The prediction results of HFS-enhanced NO are improved compared to NO for all time-steps. However, the enhancement is more pronounced at later time-steps, where the NO predictions are significantly over-smoothed. 

The average energy for the high-frequency component of the latent features (e.g., excluding the first 12.5\% frequencies at the full resolution) is generally higher for HFS-enhanced NO. This behavior is specifically seen in all the encoder layers and the last three layers of the decoder for a five-layer decoder and five-layer encoder (e.g., five downsampling and five upsampling steps). The first two layers after the bottleneck are at very low spatial resolutions and may not represent any useful spectral information. However, more high-frequency component is generated in the later stages of the decoder that are closer to the output. The NO decoder mean feature maps at each layer show low-contrast regions at both left and right side of the maps, starting from layer two to the end. However, these regions are diminished when HFS is used, showing that a more diverse set of features is generated in the decoder (see \ref{appE}). However, the same behavior does not necessarily exist for the encoder mean latent features, suggesting that the mean feature map may not be a good representative of the high-frequency component. Instead, analysis of individual feature maps appears to be a more appropriate approach in this case.\\

\begin{figure}[H]
    \centering
    \includegraphics[width=1\textwidth]{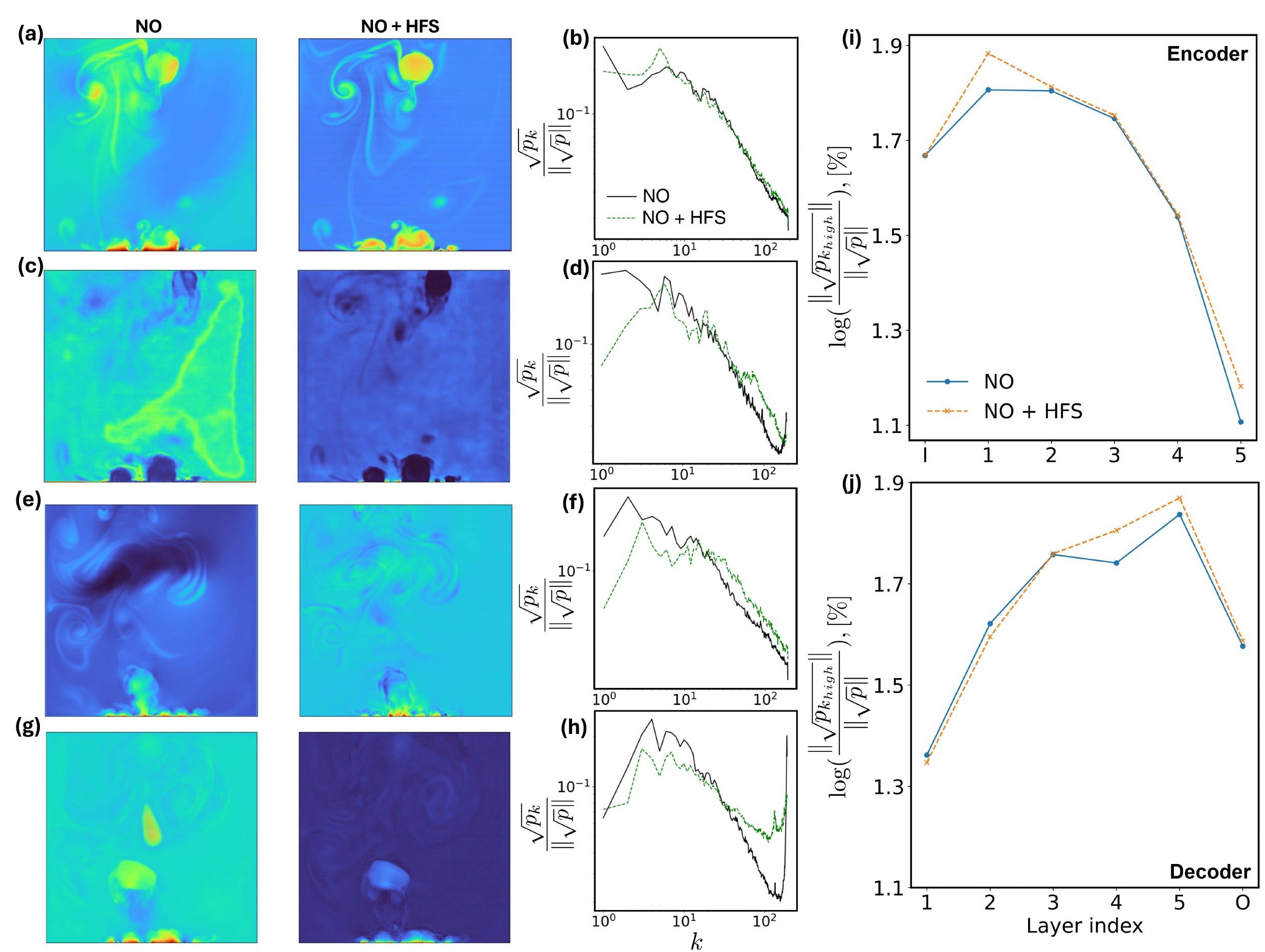}
    \caption{\textbf{Latent space features in HFS-enhanced NO.} (a, b) Example of latent space feature in the first layer of encoder and the corresponding normalized energy spectra (\(p(k)\)) in the \(\sim\) 3.5 million parameter models. (c, d) Example of latent feature in the last layer of decoder and the corresponding normalized energy spectra for the model with \(\sim\) 3.5 million parameters. (e) Example of latent feature in the first layer of encoder and the corresponding normalized energy spectra in the \(\sim\) 16 million parameter models. (g) Example of latent feature in the last layer of decoder and the corresponding normalized energy spectra in the \(\sim\) 16 million parameter models. (i-j) Average ratio of high-
    frequency energy to total energy at each layer in encoder (i) and decoder (j). Note that the low-frequency cutoff is set to the first 12.5 \%, 18.75 \%, 25 \%, 37.5 \%, and 50 \% of the wavenumbers, from highest to lowest spatial resolutions (384 to 24 pixels), respectively}
    \label{Figure 7}
\end{figure}

Individual latent space features exhibit improved preservation and propagation of high-frequency components when HFS is integrated in the NO structure. Fig. \ref{Figure 7} depicts examples of latent features from the first layer of the encoder and the last layer of decoder. These layers are specifically chosen due to their proximity to the input and output layers, making the visualizations more understandable. When comparing similar latent feature maps, HFS reduces the excessive smoothing and increases the high-frequency component within the features in the latent space. The energy spectra plots in Fig. \ref{Figure 7} demonstrate similar trends for both NO and HFS-enhanced NO with the later having larger spectral energy at the mid and high wave numbers (e.g. $k > 20$). For a more robust spectral analysis of latent features, we compared the individual latent features in the NO and HFS-enhanced NO with both \(\sim\) 3.5 and \(\sim\) 16 million parameter models. The HFS-enhanced NO decreases the over-smoothing in latent features when compared with a similar feature map from NO. The normalized energy spectra of these latent features exhibit larger high-frequency component with HFS-enhanced NO. This is evident in Fig. \ref{Figure 7}(b, d, f, and h), where the HFS-enhanced NO curves surpass the NO curves after a certain wave number. \\

Comparison of the ratio of high-frequency component energy when calculated separately for each latent feature and then averaged over all the features at each layer in the encoder also shows consistently higher values when HFS is used. The same trend is also observed in the last three layers of the decoder. These results are shown in Fig. \ref{Figure 7}i and \ref{Figure 7}j. We observed similar trends for other samples where the ratio of high-frequency component energy to total energy in the latent space is higher when HFS is integrated with the NO. However, this advantage may not be noticeable using the mean latent feature visualization at each layer. Note that for the analysis presented in Fig. \ref {Figure 7}i and \ref {Figure 7}j, we progressively increased the threshold (from 12.5\% to 50\%) for separating the low and high-frequency bands as the spatial dimension in the latent space decreases. This result is based on a random sample from the test dataset. Similar results were obtained with other samples. It should be noted that a one-to-one comparison of similar feature maps may provide a more reliable assessment, as not all feature maps carry equally significant information and some might be irrelevant for our analysis. \\

In general, the HFS-enhanced NO contains more high-frequency component in the latent space, which can help with the propagation of high-frequency information to the output, helping with the better capture of high-frequency features. The enhancement in high-frequency component is achieved without any degradation in the low-frequency components. Therefore, both field errors such as RMSE, and the spectral errors are improved (see \ref{appC}). \\

\begin{figure}[H]
    \centering
    \includegraphics[width=1\textwidth]{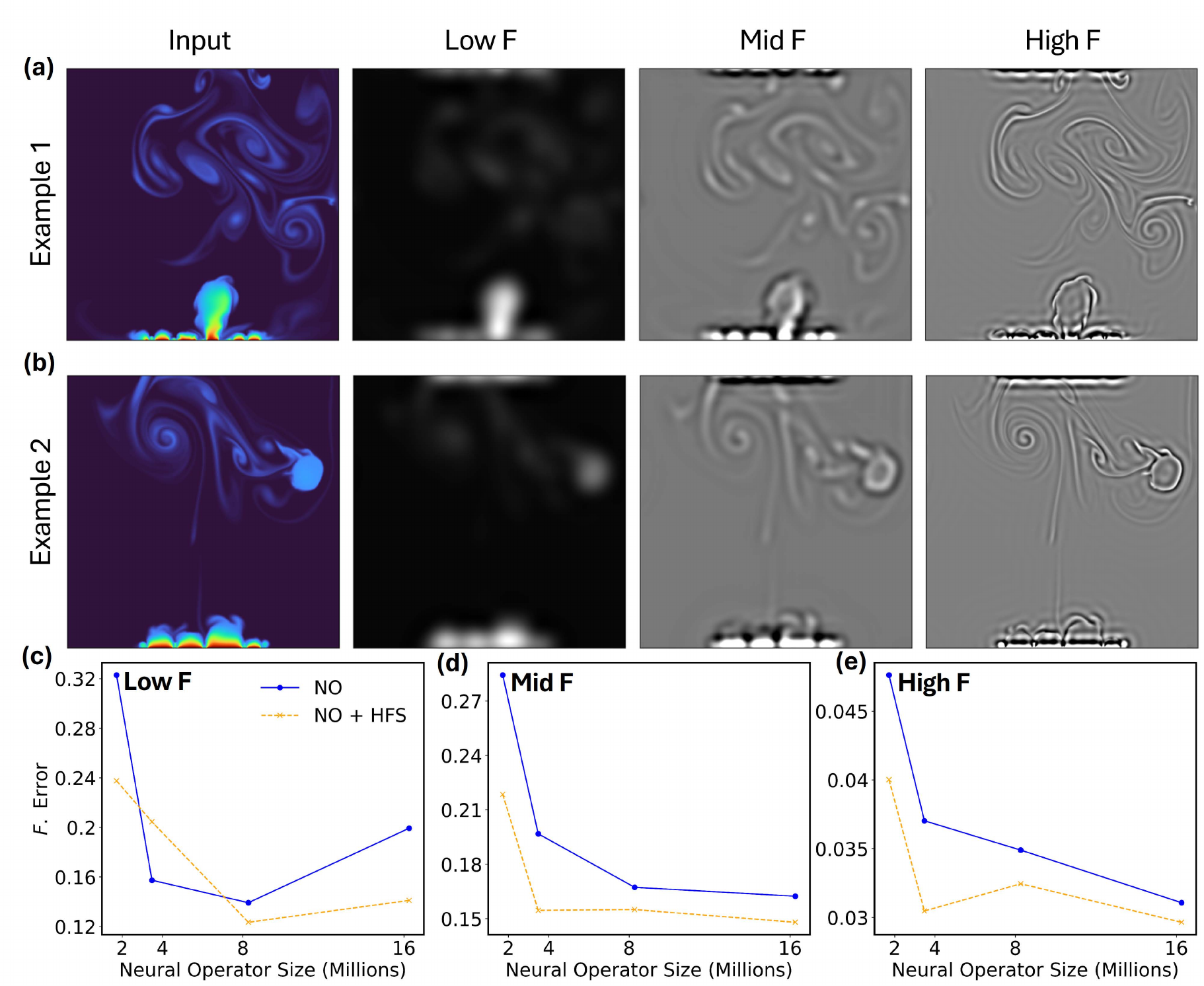}
    \caption{\textbf{Impact of HFS on spectral errors at different frequency bands} (a-b) Examples showing the input, low, mid, and high-frequency contents of the input. (c-e) Spectral error (\textit{F}. Error) of low, mid, and high-frequency bands over the test dataset. For these results, the low-frequency cutoff is set to the first 2\% of the frequencies. The mid frequency band includes the first 6.2\% of the frequencies excluding the first 2\%. The high-frequency band includes the last 93.8\% of the frequencies.}
    \label{Figure 8}
\end{figure}

Given the advantage of HFS in the mitigation of spectral bias towards low-frequency components, it is natural to calculate the prediction errors at different wavenumbers. Following the terminology proposed in \citep{hassan2023bubbleml}, we divided the frequencies to three components including only low-frequency component (low \textit{F}), mid-frequency component (mid \textit{F}), and high-frequency component (high \textit{F}). For all the NOs with varying number of parameters, the errors in the mid \textit{F} and high \textit{F} components are always lower for HFS-enhanced NO. The RMSE for the low \textit{F} component is lower for HFS-enhanced NO with one exception in the NO with \(\sim\) 3.5 million parameters. We attribute this to the larger enhancement observed in mid \textit{F} and high \textit{F} of the 3.5 million parameter HFS-enhanced NO, causing the operator showing larger error in the low \textit{F} as it fails to reduce the errors in all three components simultaneously. Visualization of each frequency component and the average spectral errors in each component are shown in Fig. \ref{Figure 8}.\\

\subsection{HFS parameter history}

\begin{figure}[H]
    \centering
    \includegraphics[width=1\textwidth]{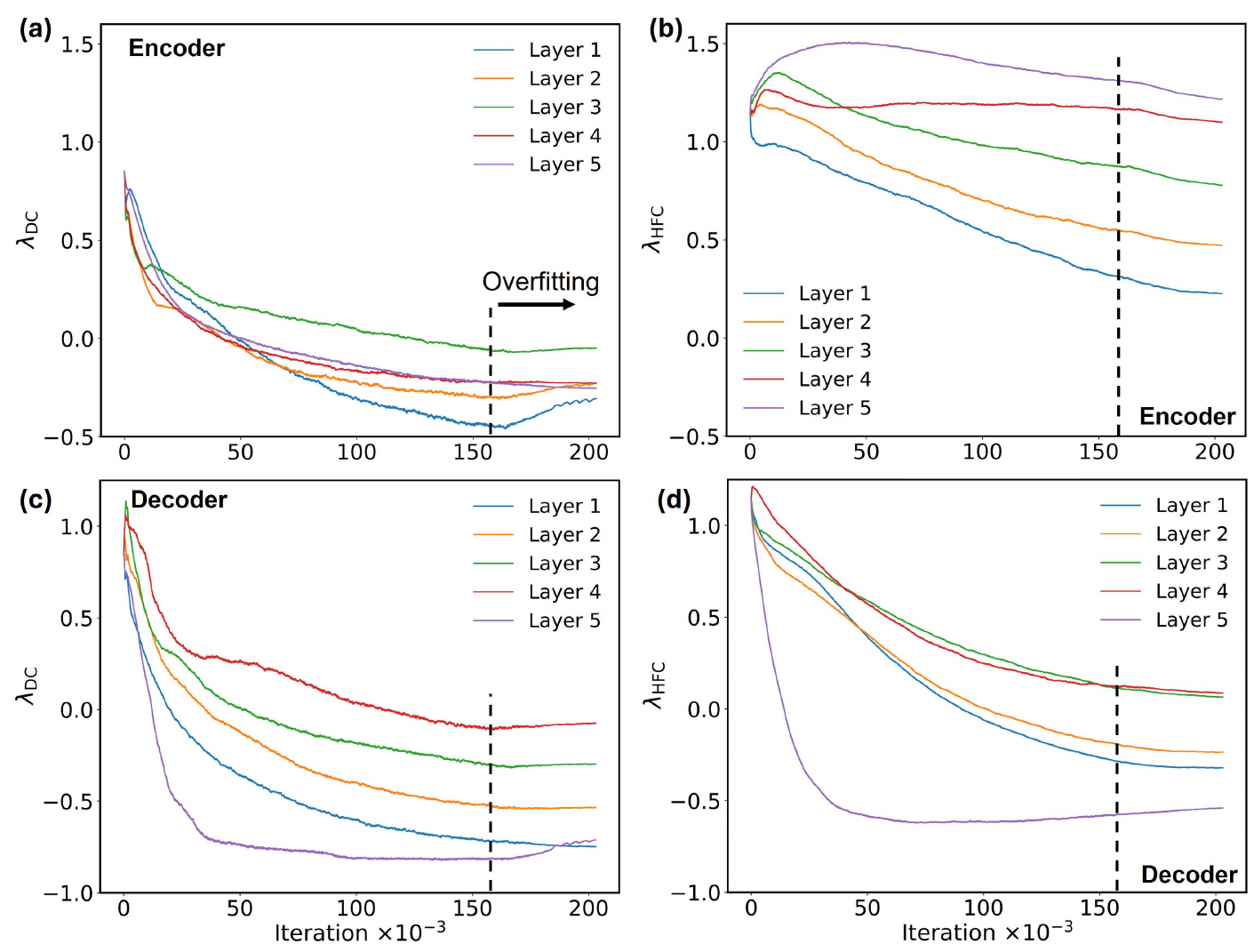}
    \caption{\textbf{\(\lambda_{DC}\) and \(\lambda_{HFC}\) histories during training phase of the HFS-enhanced NO.} (a, b) \(\lambda_{DC}\) and \(\lambda_{HFC}\) training history in all 5 layers of encoder. Note that layer 1 and layer 5 are defined as layers at highest and lowest spatial resolution, respectively, in the encoder. (c, d) \(\lambda_{DC}\) and \(\lambda_{HFC}\) training history in all 5 layers of decoder. Note that layer 1 and layer 5 are defined as layers at lowest and highest spatial resolution, respectively, in the decoder, which is the opposite terminology used in encoder. The dashed lines specify the iteration from which overfitting on the training dataset starts. The results are based on the training of a model with \(\sim\) 1.7 million parameters and \(\lambda_{DC}\) and \(\lambda_{HFC}\) were initialized at 0.85 and 1.15, respectively.}
    \label{Figure 9}
\end{figure}

The DC and HFC of the signals are scaled using two learnable parameters, \(\lambda_{DC} \in \mathbb{R}^{1 \times 1 \times C}\) and \(\lambda_{HFC} \in \mathbb{R}^{1 \times 1 \times C}\). These parameters remain consistent across all patches in each latent space feature map, and also across all batches of the dataset. Therefore, the parameters are optimized based on all the samples in the training dataset. However, they are allowed to vary freely across the feature channels at each layer. This design enables the model to adaptively scale each channel based on its content. For instance, a feature channel with a larger high-frequency component can be scaled differently than a smoother feature channel. This flexibility enhances the effectiveness of HFS while minimizing the computational costs and reducing the optimization burden by maintaining fixed parameters across patches and samples. To better understand the learning process of \(\lambda_{DC}\)  and \(\lambda_{HFC}\), the histories of these parameters during the training phase in each of the encoder and decoder layers are shown in Fig. \ref{Figure 9}. The results in Fig. \ref{Figure 9} show the mean \(\lambda_{DC}\)  and \(\lambda_{HFC}\) across all latent features at each layer. The mean \(\lambda_{HFC}\) is always larger than the mean \(\lambda_{DC}\), demonstrating that the model is learning to scale HFC with larger weights, enhancing the representation of the HFC. Also, the optimized mean \(\lambda_{HFC}\) is higher in the deeper layers of the encoder. However, no such behavior is observed in the decoder. Another interesting observation is that the abrupt change in the slope of the \(\lambda_{DC}\) history curves (\(\sim\) $\text{iteration 160} \times 10^3$) aligns well with the iteration when overfitting starts. After this iteration, the error over training dataset keeps decreasing but the error over validation dataset increases, leading to larger generalization gap. The dashed lines in Fig. \ref{Figure 9} specify the iteration at which the validation dataset error is minimum. \\

It should be noted that \(\lambda_{DC}\) and \(\lambda_{HFC}\) are both free of any constraints and are automatically learned during the model optimization. However, comparing the final values of these parameters align well with the heuristic viewpoint proposed in our work. The larger values of \(\lambda_{HFC}\) imply that the HFC of the signals are better preserved and propagated through layers with HFS. This could explain why the HFS-enhanced NO results resolve high-frequency features better, and why the spectral bias of the NO is mitigated.

\subsection{Kolmogorov flow}

\begin{figure}[H]
    \centering
    \includegraphics[width=1\textwidth]{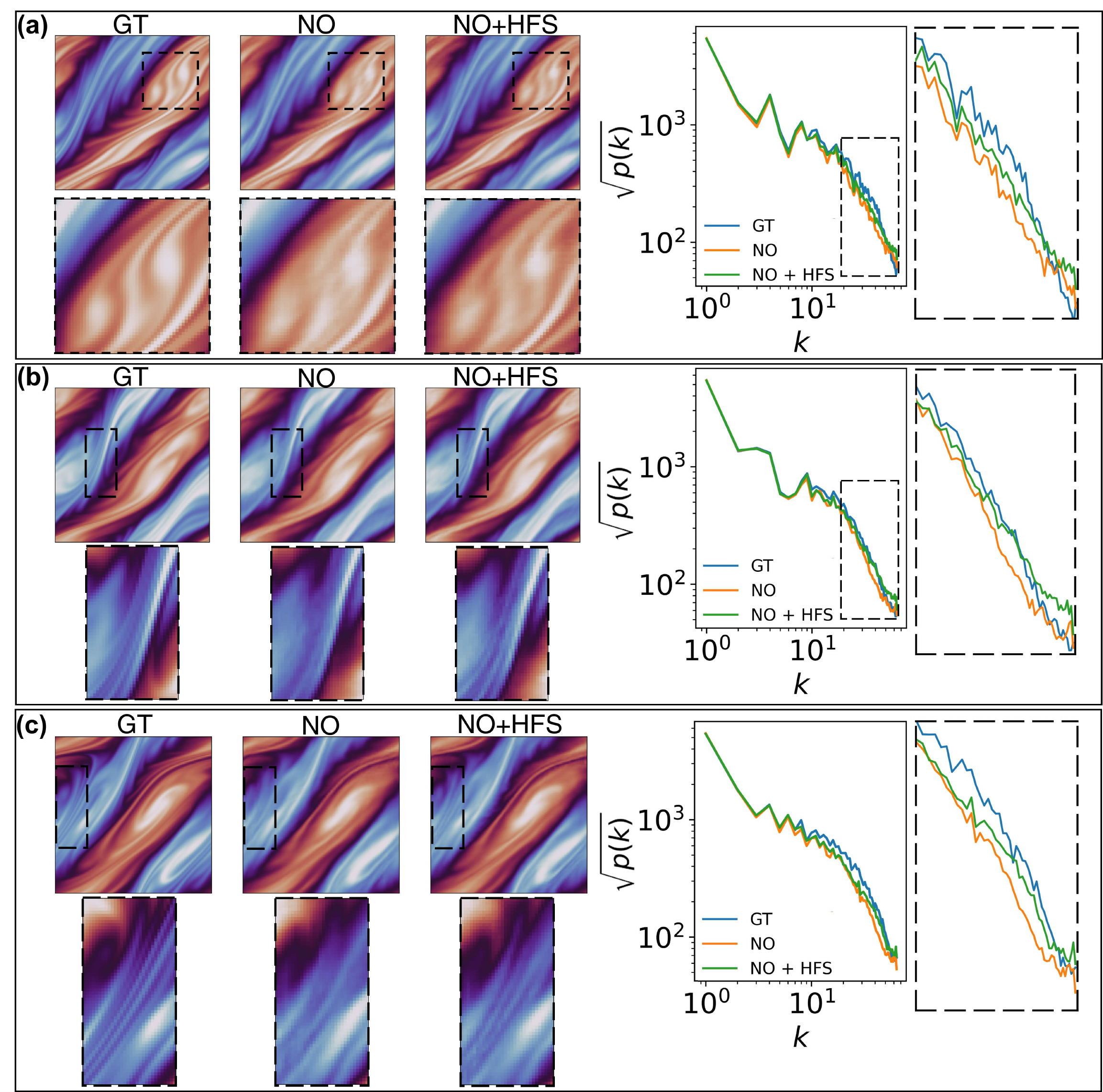}
    \caption{HFS-enhanced Kolmogorov flow predictions. (a-c) denote different samples chosen randomly from the test dataset. Each example shows the ground truth (GT), NO and HFS-enhanced NO predictions along with the energy spectra (\(p(k)\)) for each prediction.}
    \label{Figure 10}
\end{figure}

To evaluate the effectiveness of HFS in mitigating spectral bias in a more chaotic system, we applied it on the prediction of a standard benchmark, namely the 2D Kolmogorov flow problem. This problem is governed by the unsteady and incompressible Navier-Stokes equations for a viscous fluid subject to a forcing term. The vorticity form of the problem is defined in \ref{appH}. We generated the dataset \citep{Kolmogorov2025Data} using a publicly available pseudo-spectral solver \citep{li2020fourier}. The dataset consisted of 1000 samples with 80\%, 10\%, and 10\% of them being used for training, validation, and testing respectively. We trained the NO with and without HFS to learn the mapping \(\omega(x, y, t) \big|_{t \in [0,10]} \rightarrow \omega(x, y, t) \big|_{t \in [10, t_{\text{final}}]}\), where \(\omega\) is the vorticity. Here, we used \(t_{final} \) = 12.5 s, and a NO with \(\sim\) 1.7 millions parameters as the benchmark. We optimized the hyperparameters based on the NO performance without HFS and then used the same hyperparameters for training the NO with HFS. This ensured that any improvement achieved with HFS was solely attributed to its effect and not simply due to differences in optimization strategies or hyperparameters. Although not specifically designed for turbulent problems, the HFS-enhanced NO demonstrated improvements over the NO for the 2D Kolmogorov problem, reducing the relative error from 5.3\% to 4.7\%. Comparison of the energy spectra of the HFS-enhanced NO predictions also demonstrated better alignment with the ground truth solutions at high wavenumbers. The prediction results for snapshots chosen through random sampling from the test dataset are shown in Fig. \ref{Figure 10}. High-frequency features are more accurately captured and the energy spectra alignment at high wavenumbers is enhanced with the HFS-enhanced NO. We should acknowledge that HFS was effective for this problem only when the NO already provided reasonably accurate predictions. If the NO produced extremely over-smoothed predictions, integrating HFS offered little to no improvement. More detailed results showing the temporal predictions are shown in \ref{appH}. \\

The improvements in predicting Komogorov flow are less pronounced compared to the two-phase flow problem. This is due to the different underlying structures of the solution maps. The HFS approach operates by decomposing the feature maps into low-frequency and high-frequency components through observing the patches as different signals. This approach is most effective for the data with localized features, making the DC and HFC of the signals significantly different. For example, this is true for the subcooled pool boiling dataset with localized features at the bubble interface and condensation trails. For the data with similar features across all regions, the distinction between DC and HFC diminishes, thus reducing the impact of HFS.

\subsection{Diffusion Model Results}
\label{subsection3_3}
We investigated further mitigation of spectral bias using the score-based diffusion model (DM) with HFS-enhanced NO predictions as the prior. Specifically, we first conducted a systematic study to investigate the effect of NO prediction accuracy, obtained by varying the number of parameters in the NO, on the diffusion model performance. Second, we demonstrated that using HFS-enhanced NO can further help the diffusion model to match the correct energy spectra of the solutions without degrading the mean prediction errors. Since the NO predictions are used as priors to diffusion model, the accuracy of diffusion model predictions is strongly influenced by the reliability of these priors. For example, if the prior information is significantly erroneous or over-smoothed, then the diffusion model struggles to accurately recover the missing frequencies without compromising the accuracy of the mean predictions.\\

\begin{figure}[H]
    \centering
    \includegraphics[width=1\textwidth]{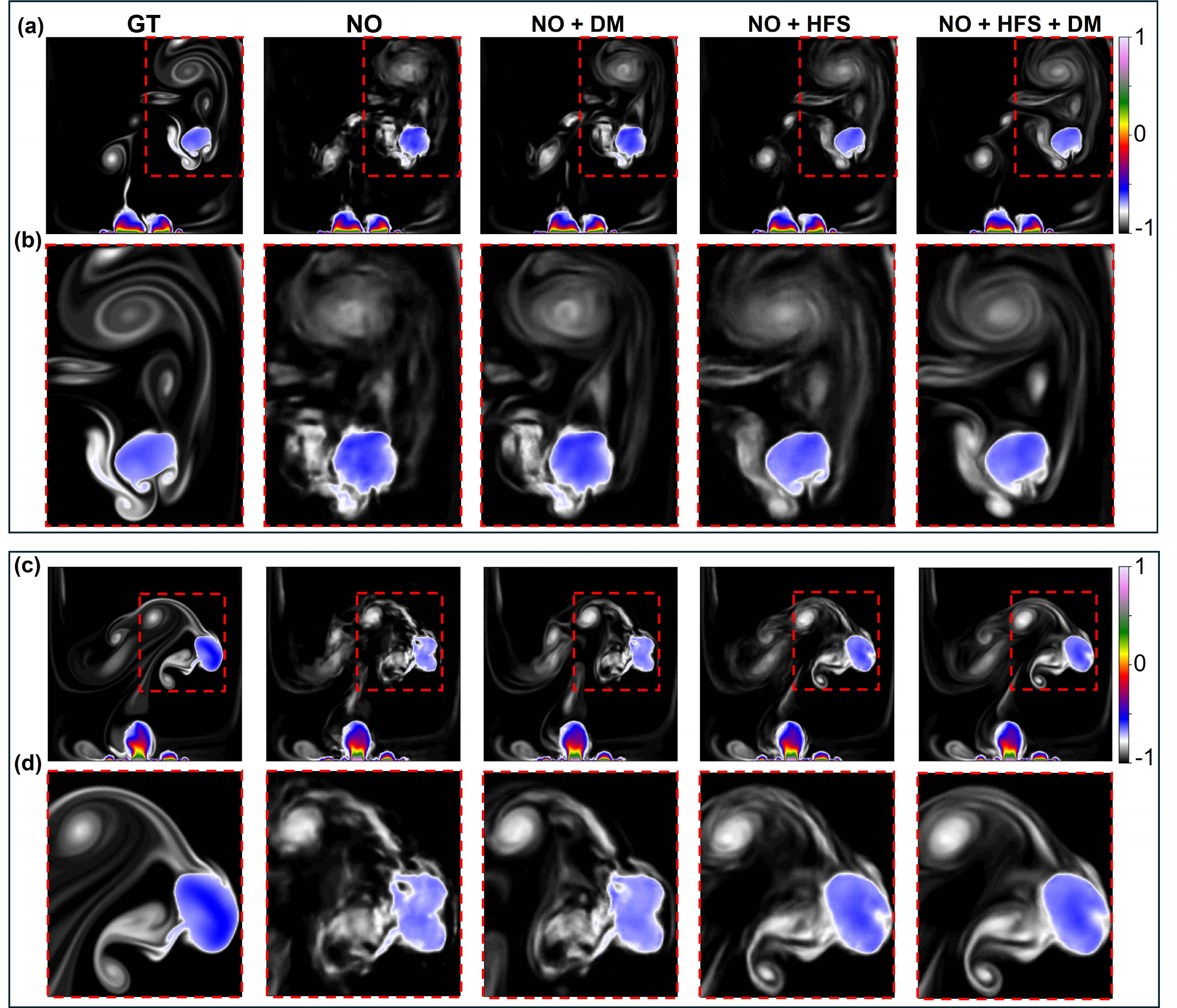}
    \caption{\textbf{Visualization of the prediction by DM integrated with NO and HFS-enhanced NO.} (a) Example showing ground truth (GT) solution and predictions by NO, NO + DM, NO + HFS, and NO + HFS + DM. (b) Zoomed-in visualization of (a) focusing on the high-frequency contents. (c) Predictions of another randomly selected sample. (d) Zoomed-in visualization of (c) focusing on high-frequency contents.}
    \label{Figure 11}
\end{figure}

\begin{figure}[H]
    \centering
    \includegraphics[width=1\textwidth]{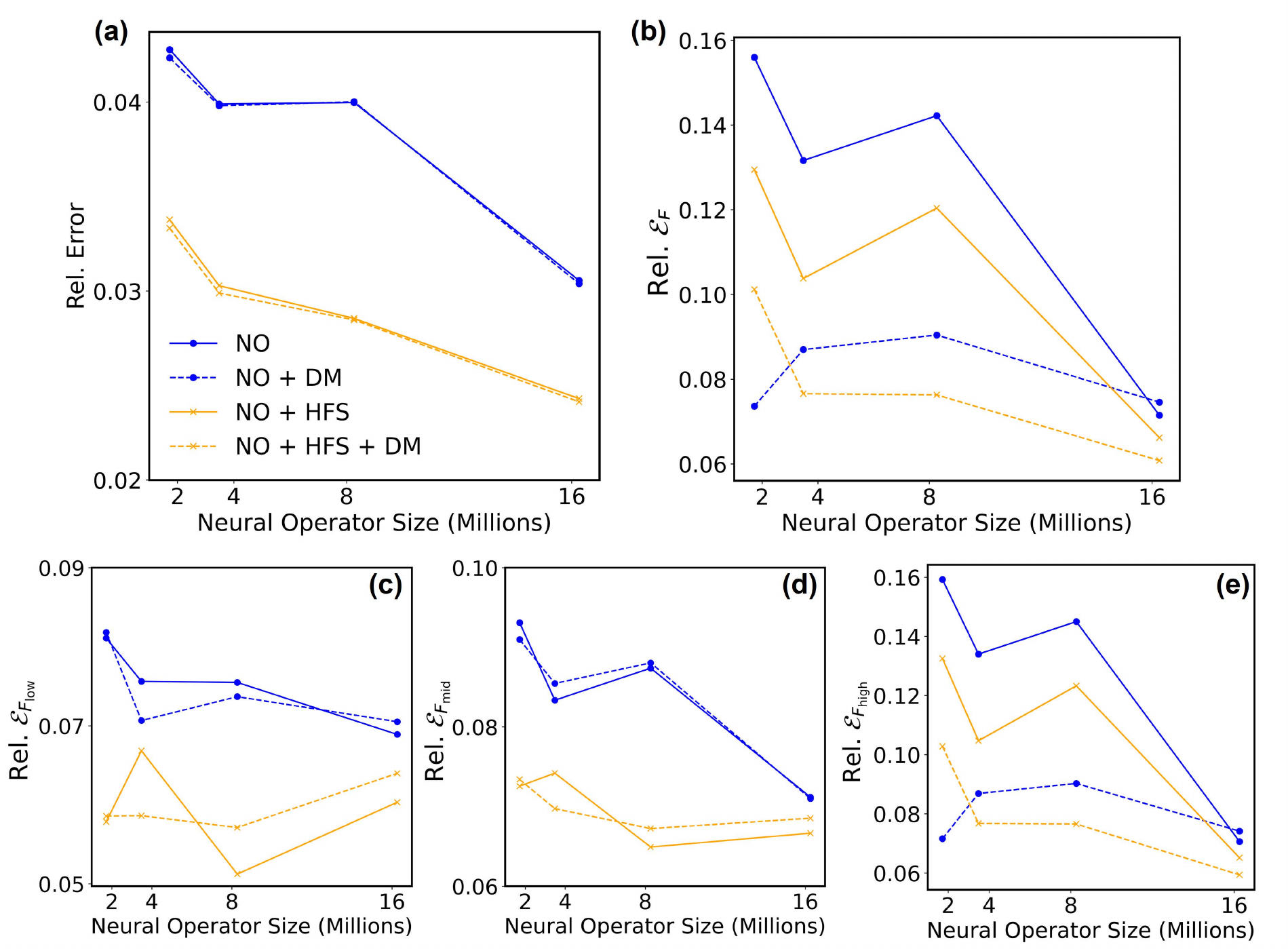}
    \caption{\textbf{Diffusion model prediction results.} (a) Relative errors (Rel. Error) of prediction by NO, NO + DM, NO + HFS, and NO + HFS + DM. (b) Relative energy spectrum errors (Rel. \(\mathcal{E}_\textit{F}\)). (c) Relative energy spectrum errors in the low frequency band (Rel.\ $\mathcal{E}_{F_{\text{low}}}$). (d) Relative energy spectrum errors in the mid frequency band (Rel.\ $\mathcal{E}_{F_{\text{mid}}}$). (e) Relative energy spectrum errors in the high-frequency band (Rel.\ $\mathcal{E}_{F_{\text{high}}}$). Low, mid, and high-frequency thresholds are set to the first 2\%, the first 6.2\% excluding the first 2\%, and the last 93.8\% of the wavenumbers.}
    \label{Figure 12}
\end{figure}

Fig. \ref{Figure 11} shows the subcooled pool boiling prediction results of DM conditioned on NO and HFS-enhanced NO predictions. Other prediction examples with DM integrated with NO and HFS-enhanced NO are visualized in Appendix F. When the NO predictions have significant errors, the DM can barely mitigate those errors. However, when HFS is integrated with the NO, the significant errors at large structures are reduced, and high-frequency components of the solutions are captured more accurately compared to NO + DM predictions. In addition, when the DM is integrated with the HFS-enhanced NO predictions, the DM is able to more accurately reconstruct intricate features that are already enhanced through more accurate predictions provided by HFS-enhanced NO. Therefore, less over-smoothing is observed in the NO + HFS + DM predictions and spectral bias is further reduced. It can be seen that both HFS and DM are helping with the capture of high-frequency features. DM cannot fix significant errors caused by NO predictions at large scale features (e.g., bubble interfaces). However, HFS reduces the errors around large scale features in addition to enhancing the smaller scale features. When DM is integrated with HFS-enhanced NO, it further enhances the small scale features. The quantitative metrics are shown in Fig. \ref{Figure 12}. It should be noted that the models are trained with a different set of hyperparameters for the results shown in Fig. \ref{Figure 12} compared to the previous results (Fig. \ref{Figure 4}). However, HFS enhanced the prediction results of NO, irrespective of hyperparameters (either optimal or non-optimal hyperparameters), as long as the same hyperparameters are used for training both NO and HFS-enhanced NO.

The results presented in Fig. \ref{Figure 11} and Fig. \ref{Figure 12} illustrate the following key points:
\begin{itemize}
    \item HFS reduces the prediction errors in both physical and spectral domains, irrespective of NO size. On average, the relative errors (e.g, RMSE) and energy spectrum errors (\(\mathcal{E}_{F}\)) (see \ref{appB}) are reduced by 23.5\% and 15.2\%, respectively, with HFS-enhanced NOs.
    \item Generally, DM does not change the prediction field errors (Fig. \ref{Figure 12}a). However, DM reduces the energy spectrum error, showing better energy spectra alignment with the correct solutions. On average, NO + DM has 27.8\% lower relative \(\mathcal{E}_{F}\) compared to NO. The only exception is the NO with 16 millions parameters. On average, NO + HFS + DM has 23.2\% lower relative \(\mathcal{E}_{F}\) compared to NO + HFS (Fig. \ref{Figure 12}b).
    \item HFS reduces the energy spectrum errors at all different frequency bands ($\mathcal{E}_{F_{\text{low}}}$, $\mathcal{E}_{F_{\text{mid}}}$, and $\mathcal{E}_{F_{\text{high}}}$), containing only the low, mid, and high-frequency components of the solutions, respectively. We refer to Fig. \ref{Figure 8} for visualization of solutions at these frequency bands. However, DM does not enhance the results at $\mathcal{E}_{F_{\text{low}}}$ and $\mathcal{E}_{F_{\text{mid}}}$ when integrated with HFS-enhanced NO. Indeed, the results at these two frequency bands are sometimes the best for HFS-enhanced NO without DM, depending on the NO size. However, the advantage of DM is taken into action at $\mathcal{E}_{F_{\text{high}}}$ (Fig. \ref{Figure 12}e) with improved results compared to NO and HFS-enhanced NO. This explains the role of DM in further mitigation of spectral bias.
    
\end{itemize}

\section{Discussion}
\label{Discussion}
HFS works by preserving more high-frequency components in the latent space after each convolutional layer in the NO. The flexibility of learning to scale the DC and HFC of the signals allows the model to enhance the predictions in mid and high-frequency contents without any degradation in the low frequency content of the solutions. As a result, both field metrics suh as RMSE and bubble RMSE, and the spectral errors are reduced in two-phase flow predictions. The enhancements observed in HFS-enhanced NO prediction results are more pronounced in areas with larger high-frequency features such as within the bubbles and at condensation trails seen in subcooled boiling solutions. This emphasizes the role of HFS in spectral bias mitigation, which helps with better capture of intricate features and sharp gradients. Similarly, both the relative errors and spectral errors are reduced, and high-frequency features are enhanced in the Kolmogorov flow predictions.\\

The scaling parameters \(\lambda_{DC}\) and \(\lambda_{HFC}\) in the HFS method are optimized during the network training. Notably, the optimized values for \(\lambda_{HFC}\) are consistently larger than \(\lambda_{DC}\), indicating that the model is trying to pay more attention to the HFC in the latent space. This biased attention helps mitigating the persistent challenge of spectral bias in the NO. To reduce the optimization burden, the scaling parameters were consistent across all the patches but were allowed to vary across different feature maps. This flexibility enables the model to automatically adjust the scaling of the HFC of the feature map depending on its content and significance. The learned \(\lambda_{DC}\) and \(\lambda_{HFC}\) for each of the latent feature maps in the HFS-enhanced NO with \(\sim\) 1.7 million parameters are shown in Appendix G. In our work, all the scaling parameters were initialized at one and they were optimized using gradient descent with the same learning rate used for training the NO (\(\sim 8\times10^{-4}\)). It would be interesting to explore faster convergence by using different initializations  and optimization frameworks for the scaling parameters in future work.\\

Another method for spectral bias mitigation is through diffusion models conditioned on NO predictions as prior information. However, using diffusion models has two drawbacks. First, the diffusion model predictions are strongly dependent on the prior information. Therefore, it can only reduce over-smoothing from reasonably accurate NO predictions. If the NO predictions are not sufficiently accurate, then the diffusion model cannot perform well. Second, training diffusion models requires extensive computational cost as each training iteration involves \textit{n}(=32) auto-regressive denoising steps to estimate the state of the solution at each time-step. In our experiments, the diffusion model training cost is approximately 2 to 4 times higher than the NO training itself. On the other hand, the HFS method requires only a small additional computational cost and negligible additional memory for training along the NO. The number of parameters added by HFS modules varies depending on the underlying NO size. However, it is generally less than 0.1\% of the number of parameters in the NO. In our experimentation, the HFS module parameters vary between 0.018\% to 0.045\% of the number of parameters in NO, depending on the underlying NO size. Based on our experiments, the computational time for each training iteration is within 10\% to 30\% higher, depending on the NO size and the computational resource.\\

In addition to the enhancements observed in field metrics such as RMSE and bubble RMSE, our investigation revealed that HFS also helps with reducing the spectral errors. We demonstrated that matching the correct energy spectra at mid and high wavenumbers is directly correlated with capturing the complex features in the solutions. We would like to emphasize the importance of considering both field errors and correct energy spectra alignment in scientific machine learning  problems. The field analysis demonstrates the average performance of the predictions. However, the energy spectra analysis reveals useful information about the prediction accuracies at different frequencies and thereby explaining the possible spectral bias and loss of useful information near interfaces, vortices, and sharp gradient areas in two-phase flow and turbulence problems. It should be noted that the predictions with enhanced energy spectra alignment is beneficial when accompanied by improved mean field predictions (e.g., RMSE) and HFS-enhanced NO results satisfy this requirement. \\

When aiming to scale the different frequency bands of the signals, a logical alternative would be to perform the scaling directly in the frequency domain rather than the physical domain. As a comparison, we implemented and compared scaling in the frequency domain with our proposed method (HFS). In this regard, let \(\mathbf{X}^{(l)} \in \mathbb{R}^{H \times W \times C}\) be the output of \textit{l}-th convolutional layer, then the feature maps can be transferred to frequency domain using a 2D Fourier transform ($\mathcal{F}$):

\begin{equation}
    \hat{\mathbf{X}}^{(l)}(:, :, c) = \mathcal{F}\left(\mathbf{X}^{(l)}(:, :, c)\right), \quad c = 1, 2, \ldots, C,
\end{equation}
where \( \hat{\mathbf{X}}^{(l)} \in \mathbb{C}^{H \times W \times C} \) includes the Fourier-transformed feature maps. The low frequency and high-frequency component of features maps can be generated by truncating \( \hat{\mathbf{X}}^{(l)}\) at a frequency threshold of $\tau$. We name these components as \(\hat{\mathbf{X}}_{\text{low}}^{(l)}\) and \(\hat{\mathbf{X}}_{\text{high}}^{(l)}\). Each Fourier-transformed feature will be scaled separately:
\begin{equation}
    \hat{\mathbf{X}}_{\text{scaled}}^{(l)} = \lambda_{\text{low}} \odot \hat{\mathbf{X}}_{\text{low}}^{(l)} + \lambda_{\text{high}} \odot \hat{\mathbf{X}}_{\text{high}}^{(l)},
\end{equation}
where \(\lambda_{\text{low}} \in \mathbb{R}^{1 \times 1 \times C}\) and \(\lambda_{\text{high}} \in \mathbb{R}^{1 \times 1 \times C}\) are learnable parameters that are optimized simultaneously with the network training. Finally, the scaled feature map is reconstructed using the inverse Fourier transform:

\begin{equation}
    \mathbf{X}_{\text{scaled}}^{(l)}(:, :, c) = \mathcal{F}^{-1}\left(\hat{\mathbf{X}}_{\text{scaled}}^{(l)}(:, :, c)\right), \quad c = 1, 2, \ldots, C.
\end{equation}

Our preliminary results demonstrate that scaling in the frequency domain also improves the two-phase flow prediction results, thus helping with the spectral bias mitigation. However, the enhancements are lower than the proposed HFS method, while the computational cost is significantly higher. This is due to the Fourier and Fourier inverse transforms required in this method. Consequently, we did not proceed with the second method. However, it may worth investigating this method in future work. There is one hyperparameter for each of these scaling methods. For the proposed HFS method, the patch size is the hyperparameter, and for scaling in the frequency domain the truncation frequency is the hyperparameter. A comparison of the prediction errors and computation costs of the two methods with a NO with \(\sim\) 1.7 million parameters is shown in Table 3.\\

\begin{table}[H]
    \caption{Comparison of the proposed HFS (Method 1) and scaling in frequency domain (Method 2).}
    \label{Table 3}
    \centering
    \begin{tabular}{c|c|c|c}
    \hline
        & \textbf{NO}
        & \textbf{NO + Method 1} & \textbf{NO + Method 2} \\
    \hline
    \textbf{Rel. Error} &0.044 &0.033& 0.034 \\ \hline
    \textbf{RMSE} &0.043 &0.033 &0.034 \\ \hline
    \textbf{BRMSE} &0.116 &0.072 &0.076\\ \hline
    \textbf{Max\textsubscript{mean}} &1.14 &0.89 &0.92\\ \hline
    \textbf{Parameters [Millions]} &1.711& 1.712 &1.712\\ \hline
    \textbf{Iteration time (s)} &31.4 &34.5 &52.6\\
    \end{tabular}
\end{table}

\subsection{Effectiveness Criteria}
The HFS approach operates by spatially decomposing the features into several patches and scaling the common DC and individual HFC of the patches separately. Our investigation showed that HFS is mostly effective on datasets with localized features such as those in subcooled pool boiling dataset. For extremely chaotic systems with globally small-scale features, the DC and HFC cannot be directly separated from spatial patching as all the patches may contain similar frequency components. To better quantify this limitation, we directly applied HFS to the samples from three different case studies with inherently different features. The samples were chosen from the subcooled pool boiling, Kolmogorov flow , and a turbulent jet problem. We found that HFS is effective for the first two problems (with the effect being less pronounced on the later one), but is not effective for the third case. \\

The turbulent jet data is from the experimental Schlieren velocimetry of turbulent helium jet in air. More details about the dataset is available in the previous work \citep{settles2022schlieren}. We directly used the publicly available Schlieren velocimetry dataset \citep{settles2022schlieren} in the raw .\textit{tif} format . All the regions in the turbulent jet have similar small-scale features (see Fig. 11),  which are different from the more localized features in the subcooled pool boiling and less localized features in the Kolmogorov flow. We directly applied HFS to these datasets and visualized the gradient magnitude in a region with high-frequency features. Additionally, we visualized the ratio of the gradient strength on a high frequency region with and without HFS, as defined by $\frac{|\nabla{x}_{\text{HFS}}|}{|\nabla {x}_{\text{baseline}}|}$, where \(x\) is the chosen region, \(\nabla\) is the gradient operator, and baseline refers to the case without HFS. This ratio compares the selectiveness in scaling the gradient of the features. The HFS approach is effective for cases where it can selectively scale the gradients across the localized features. In contrast, HFS may not be effective if it results in a uniform gradient scaling,  as it can be seen in the sample from the turbulent jet dataset.\\

Specifically, as shown in Fig. 11, the HFS approach successfully increases the gradient strength at high frequency regions in subcooled pool boiling and Kolmogorov flow. However, it scales the gradient uniformly in the turbulent jet case. Therefore, the ratio of the gradient strength with HFS to the baseline shows a less uniform solution on the subcooled pool boiling sample, followed by the Kolmogorov flow sample. However, this ratio is almost uniform for the turbulent jet case. Selective enhancement of the gradient near the edges and high-frequency features helps with the better representation of these local regions which helps the NO to better capture the high-frequency details. Since HFS is applied in the latent space, the artifacts caused by patching are mitigated and ultimately discarded in the deeper levels of the NO.

\begin{figure}[H]
    \centering
    \includegraphics[width=1\textwidth]{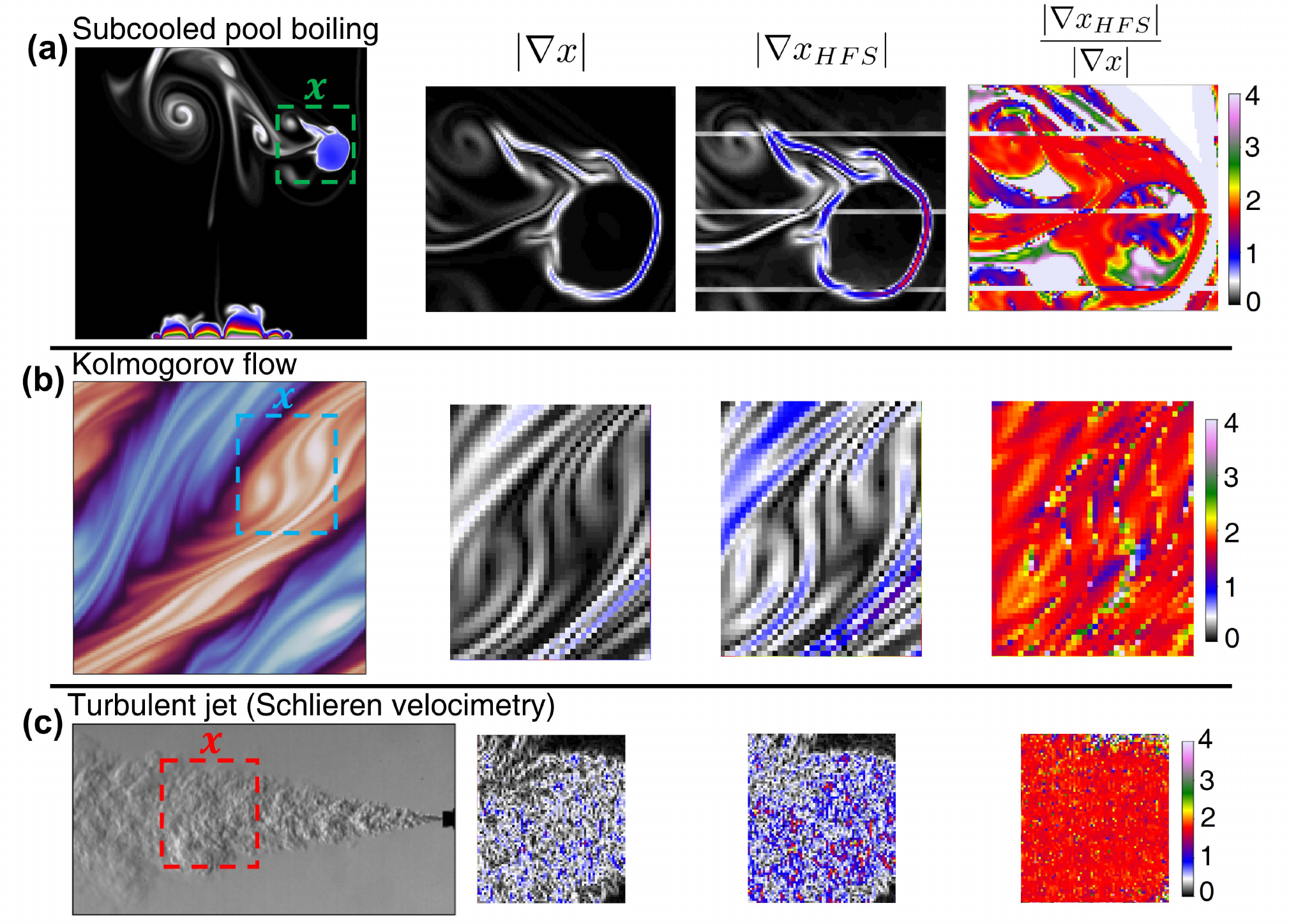}
    \caption{\textbf{HFS impact on gradient magnitude  for different problems.} (a) Subcooled pool boiling. (b) Kolmogorov flow. (c) Schlieren velocimetry of turbulent jet. For each case, the first column shows the sample and the chosen region with high frequency features (dashed boxes), the second column shows the gradient magnitude, the third column shows the gradient magnitude after applying HFS to the sample, and the fourth column  shows the ratio of the HFS-enhanced gradient magnitude to the baseline gradient magnitude.}
    \label{eff_criteria}
\end{figure}

\section{Summary}
\label{Summary}
In this work, we proposed a new method named high-frequency scaling (HFS) to mitigate the spetral bias in convolutional-based neural operators. We demonstrated that integrating HFS with feature maps in the latent space of the neural operator reduces the prediction errors in two-phase flow problems and the Kolmogorov flow problem. Through spectral bias mitigation, HFS helps to better capture intricate features and sharp gradients commonly seen within the bubbles and induced vortices in subcooled pool boiling problem, and the small-scale features in the Kolmogorov flow. These high-frequency features are prone to over-smoothing when predicted with neural operators without HFS. HFS-enhanced neural operators can improve neural operator performance irrespective of the neural operator size. We showed that for different variants of ResUNet with number of parameters varying within \( \sim\) 2 to \(\sim\) 16 millions, HFS consistently reduces the prediction errors. Furthermore, a better energy spectra alignment is observed for the results of the neural operator with HFS. Additionally, we showed that the diffusion model predictions are strongly dependent on the quality of the prior neural operator predictions. Therefore, it is important to improve the neural operator prediction accuracy using HFS so that the diffusion model can further recover the missing high-frequencies in the solutions. Otherwise, the diffusion model can barely improve the erroneous large features or significantly over-smoothed predictions of the neural operator. The advantages of HFS are obtained with a negligible memory requirement and a small computational cost trade-off.\\

Finally, we investigated the effectiveness criteria for HFS approach by visualizing the gradient magnitudes of high-frequency regions of three different problems. We showed that HFS works the best on the subcooled pool boiling dataset due to the more localized features,  which result in a selective gradient enhancement near the edges and high-frequency features. The HFS approach effectiveness decreases in the Kolmogorov flow problem,  and is negligible in the turbulent jet problem. The gradient magnitude is scaled more uniformly in the Kolmogorov flow data and almost completely uniform in the turbulent jet problem, hence explaining why HFS is ineffective for this problem.

\section*{CRediT authorship contribution statement}
\textbf{Siavash Khodakarami:} Writing - review \& editing, Writing - original draft, Visualization, Validation, Software, Methodology, Investigation, Formal analysis, Data Curation, Conceptualization.
\textbf{Vivek Oommen:} Writing - review \& editing, Writing - original draft, Visualization, Validation, Methodology, Investigation, Formal analysis, Data curation.
\textbf{Aniruddha Bora:} Writing - review \& editing, Writing - original draft, Validation, Methodology, Investigation.
\textbf{George Em Karniadakis} Writing - review \& editing, Writing - original draft, Supervision, Funding acquisition, Conceptualization.

\section*{Declaration of competing interest}
The authors declare that they have no known competing financial interests or personal relationships that could have appeared to influence the work reported in this paper.

\section*{Acknowledgments}
We would like to acknowledge funding from the Office of Naval Research as part of MURI-METHODS project with grant number N00014242545. The authors would like to acknowledge the computational resources and services at the Center for Computation and Visualization (CCV), Brown University. The experiments were also partly conducted using the Delta AI computational resources at the National Center for Supercomputing Applications at the University of Illinois Urbana-Champaign through allocation CIS240932 from the Advanced Cyberinfrastructure Coordination Ecosystem: Services \& Support (ACCESS) program, which is supported by the National Science Foundation.

\section*{Data Availability}
All codes and datasets will be made publicly available at \url{https://github.com/SiaK4/HFS_ResUNet.git} upon publication.

\bibliography{references}

\begin{thebibliography}{64}
\expandafter\ifx\csname natexlab\endcsname\relax\def\natexlab#1{#1}\fi
\providecommand{\url}[1]{\texttt{#1}}
\providecommand{\href}[2]{#2}
\providecommand{\path}[1]{#1}
\providecommand{\DOIprefix}{doi:}
\providecommand{\ArXivprefix}{arXiv:}
\providecommand{\URLprefix}{URL: }
\providecommand{\Pubmedprefix}{pmid:}
\providecommand{\doi}[1]{\href{http://dx.doi.org/#1}{\path{#1}}}
\providecommand{\Pubmed}[1]{\href{pmid:#1}{\path{#1}}}
\providecommand{\bibinfo}[2]{#2}
\ifx\xfnm\relax \def\xfnm[#1]{\unskip,\space#1}\fi
\bibitem[{Godunov and Bohachevsky(1959)}]{godunov1959finite}
\bibinfo{author}{S.~K. Godunov}, \bibinfo{author}{I.~Bohachevsky},
\newblock \bibinfo{title}{Finite difference method for numerical computation of discontinuous solutions of the equations of fluid dynamics},
\newblock \bibinfo{journal}{Matemati{\v{c}}eskij Sbornik} \bibinfo{volume}{47} (\bibinfo{year}{1959}) \bibinfo{pages}{271--306}.
\bibitem[{Eymard et~al.(2000)Eymard, Gallou{\"e}t, and Herbin}]{eymard2000finite}
\bibinfo{author}{R.~Eymard}, \bibinfo{author}{T.~Gallou{\"e}t}, \bibinfo{author}{R.~Herbin},
\newblock \bibinfo{title}{Finite volume methods},
\newblock \bibinfo{journal}{Handbook of Numerical Analysis} \bibinfo{volume}{7} (\bibinfo{year}{2000}) \bibinfo{pages}{713--1018}.
\bibitem[{Karniadakis and Sherwin(2005)}]{karniadakis2005spectral}
\bibinfo{author}{G.~Karniadakis}, \bibinfo{author}{S.~J. Sherwin}, \bibinfo{title}{Spectral/hp element methods for computational fluid dynamics}, \bibinfo{publisher}{Oxford University Press, USA}, \bibinfo{year}{2005}.
\bibitem[{Hughes(2012)}]{hughes2012finite}
\bibinfo{author}{T.~J. Hughes}, \bibinfo{title}{The Finite Element Method: Linear Static and Dynamic Finite Element Analysis}, \bibinfo{publisher}{Courier Corporation}, \bibinfo{year}{2012}.
\bibitem[{Lu et~al.(2021)Lu, Jin, Pang, Zhang, and Karniadakis}]{Lu2021}
\bibinfo{author}{L.~Lu}, \bibinfo{author}{P.~Jin}, \bibinfo{author}{G.~Pang}, \bibinfo{author}{Z.~Zhang}, \bibinfo{author}{G.~E. Karniadakis},
\newblock \bibinfo{title}{Learning nonlinear operators via deeponet based on the universal approximation theorem of operators},
\newblock \bibinfo{journal}{Nature Machine Intelligence} \bibinfo{volume}{3} (\bibinfo{year}{2021}) \bibinfo{pages}{218--229}. \URLprefix \url{https://doi.org/10.1038/s42256-021-00302-5}. \DOIprefix\doi{10.1038/s42256-021-00302-5}.
\bibitem[{Li et~al.(2020)Li, Kovachki, Azizzadenesheli, Liu, Bhattacharya, Stuart, and Anandkumar}]{li2020fourier}
\bibinfo{author}{Z.~Li}, \bibinfo{author}{N.~Kovachki}, \bibinfo{author}{K.~Azizzadenesheli}, \bibinfo{author}{B.~Liu}, \bibinfo{author}{K.~Bhattacharya}, \bibinfo{author}{A.~Stuart}, \bibinfo{author}{A.~Anandkumar},
\newblock \bibinfo{title}{Fourier neural operator for parametric partial differential equations},
\newblock \bibinfo{journal}{arXiv preprint arXiv:2010.08895}  (\bibinfo{year}{2020}).
\bibitem[{Cao et~al.(2024)Cao, Goswami, and Karniadakis}]{cao2024laplace}
\bibinfo{author}{Q.~Cao}, \bibinfo{author}{S.~Goswami}, \bibinfo{author}{G.~E. Karniadakis},
\newblock \bibinfo{title}{Laplace neural operator for solving differential equations},
\newblock \bibinfo{journal}{Nature Machine Intelligence} \bibinfo{volume}{6} (\bibinfo{year}{2024}) \bibinfo{pages}{631--640}.
\bibitem[{Tripura and Chakraborty(2023)}]{tripura2023wavelet}
\bibinfo{author}{T.~Tripura}, \bibinfo{author}{S.~Chakraborty},
\newblock \bibinfo{title}{Wavelet neural operator for solving parametric partial differential equations in computational mechanics problems},
\newblock \bibinfo{journal}{Computer Methods in Applied Mechanics and Engineering} \bibinfo{volume}{404} (\bibinfo{year}{2023}) \bibinfo{pages}{115783}.
\bibitem[{Ovadia et~al.(2023)Ovadia, Oommen, Kahana, Peyvan, Turkel, and Karniadakis}]{ovadia2023real}
\bibinfo{author}{O.~Ovadia}, \bibinfo{author}{V.~Oommen}, \bibinfo{author}{A.~Kahana}, \bibinfo{author}{A.~Peyvan}, \bibinfo{author}{E.~Turkel}, \bibinfo{author}{G.~E. Karniadakis},
\newblock \bibinfo{title}{Real-time inference and extrapolation via a diffusion-inspired temporal transformer operator (ditto)},
\newblock \bibinfo{journal}{arXiv preprint arXiv:2307.09072}  (\bibinfo{year}{2023}).
\bibitem[{Li et~al.(2022)Li, Meidani, and Farimani}]{li2022transformer}
\bibinfo{author}{Z.~Li}, \bibinfo{author}{K.~Meidani}, \bibinfo{author}{A.~B. Farimani},
\newblock \bibinfo{title}{Transformer for partial differential equations' operator learning},
\newblock \bibinfo{journal}{arXiv preprint arXiv:2205.13671}  (\bibinfo{year}{2022}).
\bibitem[{Sharma et~al.(2024)Sharma, Singh, and Ratna}]{sharma2024graph}
\bibinfo{author}{A.~Sharma}, \bibinfo{author}{S.~Singh}, \bibinfo{author}{S.~Ratna},
\newblock \bibinfo{title}{Graph neural network operators: a review},
\newblock \bibinfo{journal}{Multimedia Tools and Applications} \bibinfo{volume}{83} (\bibinfo{year}{2024}) \bibinfo{pages}{23413--23436}.
\bibitem[{Chen and Chen(1995)}]{approx_the}
\bibinfo{author}{T.~Chen}, \bibinfo{author}{H.~Chen},
\newblock \bibinfo{title}{Universal approximation to nonlinear operators by neural networks with arbitrary activation functions and its application to dynamical systems},
\newblock \bibinfo{journal}{IEEE Transactions on Neural Networks} \bibinfo{volume}{6} (\bibinfo{year}{1995}) \bibinfo{pages}{911--917}. \DOIprefix\doi{10.1109/72.392253}.
\bibitem[{Wan et~al.(2025)Wan, Kharazmi, Triantafyllou, and Karniadakis}]{wan2025deepvivonet}
\bibinfo{author}{R.~Wan}, \bibinfo{author}{E.~Kharazmi}, \bibinfo{author}{M.~S. Triantafyllou}, \bibinfo{author}{G.~E. Karniadakis},
\newblock \bibinfo{title}{Deepvivonet: Using deep neural operators to optimize sensor locations with application to vortex-induced vibrations},
\newblock \bibinfo{journal}{arXiv preprint arXiv:2501.04105}  (\bibinfo{year}{2025}).
\bibitem[{Kiyani et~al.(2024)Kiyani, Manav, Kadivar, De~Lorenzis, and Karniadakis}]{kiyani2024predicting}
\bibinfo{author}{E.~Kiyani}, \bibinfo{author}{M.~Manav}, \bibinfo{author}{N.~Kadivar}, \bibinfo{author}{L.~De~Lorenzis}, \bibinfo{author}{G.~E. Karniadakis},
\newblock \bibinfo{title}{Predicting crack nucleation and propagation in brittle materials using deep operator networks with diverse trunk architectures},
\newblock \bibinfo{journal}{arXiv preprint arXiv:2501.00016}  (\bibinfo{year}{2024}).
\bibitem[{Peyvan et~al.(2024)Peyvan, Oommen, Jagtap, and Karniadakis}]{peyvan2024riemannonets}
\bibinfo{author}{A.~Peyvan}, \bibinfo{author}{V.~Oommen}, \bibinfo{author}{A.~D. Jagtap}, \bibinfo{author}{G.~E. Karniadakis},
\newblock \bibinfo{title}{Riemannonets: Interpretable neural operators for riemann problems},
\newblock \bibinfo{journal}{Computer Methods in Applied Mechanics and Engineering} \bibinfo{volume}{426} (\bibinfo{year}{2024}) \bibinfo{pages}{116996}.
\bibitem[{Li et~al.(2023)Li, Peng, Yuan, and Wang}]{li2023long}
\bibinfo{author}{Z.~Li}, \bibinfo{author}{W.~Peng}, \bibinfo{author}{Z.~Yuan}, \bibinfo{author}{J.~Wang},
\newblock \bibinfo{title}{Long-term predictions of turbulence by implicit u-net enhanced fourier neural operator},
\newblock \bibinfo{journal}{Physics of Fluids} \bibinfo{volume}{35} (\bibinfo{year}{2023}).
\bibitem[{Jiang et~al.(2025)Jiang, Li, Wang, Yang, and Wang}]{jiang2025implicit}
\bibinfo{author}{Y.~Jiang}, \bibinfo{author}{Z.~Li}, \bibinfo{author}{Y.~Wang}, \bibinfo{author}{H.~Yang}, \bibinfo{author}{J.~Wang},
\newblock \bibinfo{title}{An implicit adaptive fourier neural operator for long-term predictions of three-dimensional turbulence},
\newblock \bibinfo{journal}{arXiv preprint arXiv:2501.12740}  (\bibinfo{year}{2025}).
\bibitem[{Gopakumar et~al.(2023)Gopakumar, Pamela, Zanisi, Li, Anandkumar, and Team}]{plasma}
\bibinfo{author}{V.~Gopakumar}, \bibinfo{author}{S.~Pamela}, \bibinfo{author}{L.~Zanisi}, \bibinfo{author}{Z.~Li}, \bibinfo{author}{A.~Anandkumar}, \bibinfo{author}{M.~Team},
\newblock \bibinfo{title}{Fourier neural operator for plasma modelling},
\newblock \bibinfo{journal}{arXiv preprint arXiv:2302.06542}  (\bibinfo{year}{2023}).
\bibitem[{Montes~de Oca~Zapiain et~al.(2021)Montes~de Oca~Zapiain, Stewart, and Dingreville}]{montes2021accelerating}
\bibinfo{author}{D.~Montes~de Oca~Zapiain}, \bibinfo{author}{J.~A. Stewart}, \bibinfo{author}{R.~Dingreville},
\newblock \bibinfo{title}{Accelerating phase-field-based microstructure evolution predictions via surrogate models trained by machine learning methods},
\newblock \bibinfo{journal}{npj Computational Materials} \bibinfo{volume}{7} (\bibinfo{year}{2021}) \bibinfo{pages}{3}.
\bibitem[{Oommen et~al.(2022)Oommen, Shukla, Goswami, Dingreville, and Karniadakis}]{oommen2022learning}
\bibinfo{author}{V.~Oommen}, \bibinfo{author}{K.~Shukla}, \bibinfo{author}{S.~Goswami}, \bibinfo{author}{R.~Dingreville}, \bibinfo{author}{G.~E. Karniadakis},
\newblock \bibinfo{title}{Learning two-phase microstructure evolution using neural operators and autoencoder architectures},
\newblock \bibinfo{journal}{npj Computational Materials} \bibinfo{volume}{8} (\bibinfo{year}{2022}) \bibinfo{pages}{190}.
\bibitem[{Oommen et~al.(2024)Oommen, Shukla, Desai, Dingreville, and Karniadakis}]{oommen2024rethinking}
\bibinfo{author}{V.~Oommen}, \bibinfo{author}{K.~Shukla}, \bibinfo{author}{S.~Desai}, \bibinfo{author}{R.~Dingreville}, \bibinfo{author}{G.~E. Karniadakis},
\newblock \bibinfo{title}{Rethinking materials simulations: Blending direct numerical simulations with neural operators},
\newblock \bibinfo{journal}{npj Computational Materials} \bibinfo{volume}{10} (\bibinfo{year}{2024}) \bibinfo{pages}{145}.
\bibitem[{Khodakarami et~al.(2023)Khodakarami, Suh, Won, and Miljkovic}]{khodakarami2023intelligent}
\bibinfo{author}{S.~Khodakarami}, \bibinfo{author}{Y.~Suh}, \bibinfo{author}{Y.~Won}, \bibinfo{author}{N.~Miljkovic},
\newblock \bibinfo{title}{An intelligent strategy for phase change heat and mass transfer: Application of machine learning},
\newblock in: \bibinfo{booktitle}{Advances in Heat Transfer}, volume~\bibinfo{volume}{56}, \bibinfo{publisher}{Elsevier}, \bibinfo{year}{2023}, pp. \bibinfo{pages}{113--168}.
\bibitem[{Rahaman et~al.(2019)Rahaman, Baratin, Arpit, Draxler, Lin, Hamprecht, Bengio, and Courville}]{spectral_bias}
\bibinfo{author}{N.~Rahaman}, \bibinfo{author}{A.~Baratin}, \bibinfo{author}{D.~Arpit}, \bibinfo{author}{F.~Draxler}, \bibinfo{author}{M.~Lin}, \bibinfo{author}{F.~Hamprecht}, \bibinfo{author}{Y.~Bengio}, \bibinfo{author}{A.~Courville},
\newblock \bibinfo{title}{On the spectral bias of neural networks},
\newblock in: \bibinfo{editor}{K.~Chaudhuri}, \bibinfo{editor}{R.~Salakhutdinov} (Eds.), \bibinfo{booktitle}{Proceedings of the 36th International Conference on Machine Learning}, volume~\bibinfo{volume}{97} of \textit{\bibinfo{series}{Proceedings of Machine Learning Research}}, \bibinfo{publisher}{PMLR}, \bibinfo{year}{2019}, pp. \bibinfo{pages}{5301--5310}. \URLprefix \url{https://proceedings.mlr.press/v97/rahaman19a.html}.
\bibitem[{Xu et~al.(2019)Xu, Zhang, Luo, Xiao, and Ma}]{spectral2}
\bibinfo{author}{Z.-Q.~J. Xu}, \bibinfo{author}{Y.~Zhang}, \bibinfo{author}{T.~Luo}, \bibinfo{author}{Y.~Xiao}, \bibinfo{author}{Z.~Ma},
\newblock \bibinfo{title}{Frequency principle: Fourier analysis sheds light on deep neural networks},
\newblock \bibinfo{journal}{arXiv preprint arXiv:1901.06523}  (\bibinfo{year}{2019}).
\bibitem[{Xu et~al.(2025)Xu, Zhang, and Cai}]{xu2025understanding}
\bibinfo{author}{Z.-Q.~J. Xu}, \bibinfo{author}{L.~Zhang}, \bibinfo{author}{W.~Cai},
\newblock \bibinfo{title}{On understanding and overcoming spectral biases of deep neural network learning methods for solving pdes},
\newblock \bibinfo{journal}{arXiv preprint arXiv:2501.09987}  (\bibinfo{year}{2025}).
\bibitem[{Lin et~al.(2021)Lin, Li, Lu, Cai, Maxey, and Karniadakis}]{bubble_DeepONet}
\bibinfo{author}{C.~Lin}, \bibinfo{author}{Z.~Li}, \bibinfo{author}{L.~Lu}, \bibinfo{author}{S.~Cai}, \bibinfo{author}{M.~Maxey}, \bibinfo{author}{G.~E. Karniadakis},
\newblock \bibinfo{title}{Operator learning for predicting multiscale bubble growth dynamics},
\newblock \bibinfo{journal}{The Journal of Chemical Physics} \bibinfo{volume}{154} (\bibinfo{year}{2021}).
\bibitem[{Jain et~al.(2025)Jain, Roy, Kodamana, and Nair}]{mp_flow_simple}
\bibinfo{author}{N.~Jain}, \bibinfo{author}{S.~Roy}, \bibinfo{author}{H.~Kodamana}, \bibinfo{author}{P.~Nair},
\newblock \bibinfo{title}{Scaling the predictions of multiphase flow through porous media using operator learning},
\newblock \bibinfo{journal}{Chemical Engineering Journal} \bibinfo{volume}{503} (\bibinfo{year}{2025}) \bibinfo{pages}{157671}.
\bibitem[{Ronneberger et~al.(2015)Ronneberger, Fischer, and Brox}]{ronneberger2015u}
\bibinfo{author}{O.~Ronneberger}, \bibinfo{author}{P.~Fischer}, \bibinfo{author}{T.~Brox},
\newblock \bibinfo{title}{U-net: Convolutional networks for biomedical image segmentation},
\newblock in: \bibinfo{booktitle}{Medical image computing and computer-assisted intervention--MICCAI 2015: 18th international conference, Munich, Germany, October 5-9, 2015, proceedings, part III 18}, \bibinfo{organization}{Springer}, \bibinfo{year}{2015}, pp. \bibinfo{pages}{234--241}.
\bibitem[{Qin et~al.(2024)Qin, Lyu, Peng, Geng, Wang, Gao, Liu, and Wang}]{FNO_spectral}
\bibinfo{author}{S.~Qin}, \bibinfo{author}{F.~Lyu}, \bibinfo{author}{W.~Peng}, \bibinfo{author}{D.~Geng}, \bibinfo{author}{J.~Wang}, \bibinfo{author}{N.~Gao}, \bibinfo{author}{X.~Liu}, \bibinfo{author}{L.~L. Wang},
\newblock \bibinfo{title}{Toward a better understanding of fourier neural operators: Analysis and improvement from a spectral perspective},
\newblock \bibinfo{journal}{arXiv preprint arXiv:2404.07200}  (\bibinfo{year}{2024}).
\bibitem[{Hassan et~al.(2024)Hassan, Feeney, Dhruv, Kim, Suh, Ryu, Won, and Chandramowlishwaran}]{bubbleML}
\bibinfo{author}{S.~M.~S. Hassan}, \bibinfo{author}{A.~Feeney}, \bibinfo{author}{A.~Dhruv}, \bibinfo{author}{J.~Kim}, \bibinfo{author}{Y.~Suh}, \bibinfo{author}{J.~Ryu}, \bibinfo{author}{Y.~Won}, \bibinfo{author}{A.~Chandramowlishwaran},
\newblock \bibinfo{title}{Bubbleml: A multiphase multiphysics dataset and benchmarks for machine learning},
\newblock \bibinfo{journal}{Advances in Neural Information Processing Systems} \bibinfo{volume}{36} (\bibinfo{year}{2024}).
\bibitem[{Dubey et~al.(2022)Dubey, Weide, O’Neal, Dhruv, Couch, Harris, Klosterman, Jain, Rudi, Messer et~al.}]{FlashX}
\bibinfo{author}{A.~Dubey}, \bibinfo{author}{K.~Weide}, \bibinfo{author}{J.~O’Neal}, \bibinfo{author}{A.~Dhruv}, \bibinfo{author}{S.~Couch}, \bibinfo{author}{J.~A. Harris}, \bibinfo{author}{T.~Klosterman}, \bibinfo{author}{R.~Jain}, \bibinfo{author}{J.~Rudi}, \bibinfo{author}{B.~Messer}, et~al.,
\newblock \bibinfo{title}{Flash-x: A multiphysics simulation software instrument},
\newblock \bibinfo{journal}{SoftwareX} \bibinfo{volume}{19} (\bibinfo{year}{2022}) \bibinfo{pages}{101168}.
\bibitem[{Liu et~al.(2024)Liu, Xu, Cao, and Zhang}]{turbulence_spectral}
\bibinfo{author}{X.~Liu}, \bibinfo{author}{B.~Xu}, \bibinfo{author}{S.~Cao}, \bibinfo{author}{L.~Zhang},
\newblock \bibinfo{title}{Mitigating spectral bias for the multiscale operator learning},
\newblock \bibinfo{journal}{Journal of Computational Physics} \bibinfo{volume}{506} (\bibinfo{year}{2024}) \bibinfo{pages}{112944}.
\bibitem[{Cai and Xu(2019)}]{cai2019multi}
\bibinfo{author}{W.~Cai}, \bibinfo{author}{Z.-Q.~J. Xu},
\newblock \bibinfo{title}{Multi-scale deep neural networks for solving high dimensional pdes},
\newblock \bibinfo{journal}{arXiv preprint arXiv:1910.11710}  (\bibinfo{year}{2019}).
\bibitem[{Tancik et~al.(2020)Tancik, Srinivasan, Mildenhall, Fridovich-Keil, Raghavan, Singhal, Ramamoorthi, Barron, and Ng}]{tancik2020fourier}
\bibinfo{author}{M.~Tancik}, \bibinfo{author}{P.~Srinivasan}, \bibinfo{author}{B.~Mildenhall}, \bibinfo{author}{S.~Fridovich-Keil}, \bibinfo{author}{N.~Raghavan}, \bibinfo{author}{U.~Singhal}, \bibinfo{author}{R.~Ramamoorthi}, \bibinfo{author}{J.~Barron}, \bibinfo{author}{R.~Ng},
\newblock \bibinfo{title}{Fourier features let networks learn high frequency functions in low dimensional domains},
\newblock \bibinfo{journal}{Advances in neural information processing systems} \bibinfo{volume}{33} (\bibinfo{year}{2020}) \bibinfo{pages}{7537--7547}.
\bibitem[{Wang et~al.(2021)Wang, Wang, and Perdikaris}]{wang2021eigenvector}
\bibinfo{author}{S.~Wang}, \bibinfo{author}{H.~Wang}, \bibinfo{author}{P.~Perdikaris},
\newblock \bibinfo{title}{On the eigenvector bias of fourier feature networks: From regression to solving multi-scale pdes with physics-informed neural networks},
\newblock \bibinfo{journal}{Computer Methods in Applied Mechanics and Engineering} \bibinfo{volume}{384} (\bibinfo{year}{2021}) \bibinfo{pages}{113938}.
\bibitem[{Raissi et~al.(2019)Raissi, Perdikaris, and Karniadakis}]{raissi2019physics}
\bibinfo{author}{M.~Raissi}, \bibinfo{author}{P.~Perdikaris}, \bibinfo{author}{G.~E. Karniadakis},
\newblock \bibinfo{title}{Physics-informed neural networks: A deep learning framework for solving forward and inverse problems involving nonlinear partial differential equations},
\newblock \bibinfo{journal}{Journal of Computational physics} \bibinfo{volume}{378} (\bibinfo{year}{2019}) \bibinfo{pages}{686--707}.
\bibitem[{Toscano et~al.(2024)Toscano, Oommen, Varghese, Zou, Daryakenari, Wu, and Karniadakis}]{toscano2024pinns}
\bibinfo{author}{J.~D. Toscano}, \bibinfo{author}{V.~Oommen}, \bibinfo{author}{A.~J. Varghese}, \bibinfo{author}{Z.~Zou}, \bibinfo{author}{N.~A. Daryakenari}, \bibinfo{author}{C.~Wu}, \bibinfo{author}{G.~E. Karniadakis},
\newblock \bibinfo{title}{From pinns to pikans: Recent advances in physics-informed machine learning},
\newblock \bibinfo{journal}{arXiv preprint arXiv:2410.13228}  (\bibinfo{year}{2024}).
\bibitem[{Liang et~al.(2021)Liang, Lyu, Wang, and Yang}]{liang2021reproducing}
\bibinfo{author}{S.~Liang}, \bibinfo{author}{L.~Lyu}, \bibinfo{author}{C.~Wang}, \bibinfo{author}{H.~Yang},
\newblock \bibinfo{title}{Reproducing activation function for deep learning},
\newblock \bibinfo{journal}{arXiv preprint arXiv:2101.04844}  (\bibinfo{year}{2021}).
\bibitem[{Jagtap et~al.(2020)Jagtap, Kawaguchi, and Karniadakis}]{jagtap2020adaptive}
\bibinfo{author}{A.~D. Jagtap}, \bibinfo{author}{K.~Kawaguchi}, \bibinfo{author}{G.~E. Karniadakis},
\newblock \bibinfo{title}{Adaptive activation functions accelerate convergence in deep and physics-informed neural networks},
\newblock \bibinfo{journal}{Journal of Computational Physics} \bibinfo{volume}{404} (\bibinfo{year}{2020}) \bibinfo{pages}{109136}.
\bibitem[{Cai et~al.(2020)Cai, Li, and Liu}]{cai2020phase}
\bibinfo{author}{W.~Cai}, \bibinfo{author}{X.~Li}, \bibinfo{author}{L.~Liu},
\newblock \bibinfo{title}{A phase shift deep neural network for high frequency approximation and wave problems},
\newblock \bibinfo{journal}{SIAM Journal on Scientific Computing} \bibinfo{volume}{42} (\bibinfo{year}{2020}) \bibinfo{pages}{A3285--A3312}.
\bibitem[{Lippe et~al.(2023)Lippe, Veeling, Perdikaris, Turner, and Brandstetter}]{lippe2023pde}
\bibinfo{author}{P.~Lippe}, \bibinfo{author}{B.~Veeling}, \bibinfo{author}{P.~Perdikaris}, \bibinfo{author}{R.~Turner}, \bibinfo{author}{J.~Brandstetter},
\newblock \bibinfo{title}{Pde-refiner: Achieving accurate long rollouts with neural pde solvers},
\newblock \bibinfo{journal}{Advances in Neural Information Processing Systems} \bibinfo{volume}{36} (\bibinfo{year}{2023}) \bibinfo{pages}{67398--67433}.
\bibitem[{Zhang et~al.(2024)Zhang, Kahana, Kopani{\v{c}}{\'a}kov{\'a}, Turkel, Ranade, Pathak, and Karniadakis}]{zhang2024blending}
\bibinfo{author}{E.~Zhang}, \bibinfo{author}{A.~Kahana}, \bibinfo{author}{A.~Kopani{\v{c}}{\'a}kov{\'a}}, \bibinfo{author}{E.~Turkel}, \bibinfo{author}{R.~Ranade}, \bibinfo{author}{J.~Pathak}, \bibinfo{author}{G.~E. Karniadakis},
\newblock \bibinfo{title}{Blending neural operators and relaxation methods in pde numerical solvers},
\newblock \bibinfo{journal}{Nature Machine Intelligence}  (\bibinfo{year}{2024}) \bibinfo{pages}{1--11}.
\bibitem[{Wu et~al.(2024)Wu, Zhang, Zhou, Chen, Han, and Cao}]{wu2024high}
\bibinfo{author}{H.~Wu}, \bibinfo{author}{K.~Zhang}, \bibinfo{author}{D.~Zhou}, \bibinfo{author}{W.-L. Chen}, \bibinfo{author}{Z.~Han}, \bibinfo{author}{Y.~Cao},
\newblock \bibinfo{title}{High-flexibility reconstruction of small-scale motions in wall turbulence using a generalized zero-shot learning},
\newblock \bibinfo{journal}{Journal of Fluid Mechanics} \bibinfo{volume}{990} (\bibinfo{year}{2024}) \bibinfo{pages}{R1}.
\bibitem[{Wang et~al.(2022)Wang, Li, Liu, Wu, Hao, Zhang, and He}]{wang2022deep}
\bibinfo{author}{Z.~Wang}, \bibinfo{author}{X.~Li}, \bibinfo{author}{L.~Liu}, \bibinfo{author}{X.~Wu}, \bibinfo{author}{P.~Hao}, \bibinfo{author}{X.~Zhang}, \bibinfo{author}{F.~He},
\newblock \bibinfo{title}{Deep-learning-based super-resolution reconstruction of high-speed imaging in fluids},
\newblock \bibinfo{journal}{Physics of Fluids} \bibinfo{volume}{34} (\bibinfo{year}{2022}).
\bibitem[{Molinaro et~al.(2024)Molinaro, Lanthaler, Raoni{\'c}, Rohner, Armegioiu, Wan, Sha, Mishra, and Zepeda-N{\'u}{\~n}ez}]{molinaro2024generative}
\bibinfo{author}{R.~Molinaro}, \bibinfo{author}{S.~Lanthaler}, \bibinfo{author}{B.~Raoni{\'c}}, \bibinfo{author}{T.~Rohner}, \bibinfo{author}{V.~Armegioiu}, \bibinfo{author}{Z.~Y. Wan}, \bibinfo{author}{F.~Sha}, \bibinfo{author}{S.~Mishra}, \bibinfo{author}{L.~Zepeda-N{\'u}{\~n}ez},
\newblock \bibinfo{title}{Generative ai for fast and accurate statistical computation of fluids},
\newblock \bibinfo{journal}{arXiv preprint arXiv:2409.18359}  (\bibinfo{year}{2024}).
\bibitem[{Lockwood et~al.(2024)Lockwood, Gori, and Gentine}]{lockwood2024generative}
\bibinfo{author}{J.~W. Lockwood}, \bibinfo{author}{A.~Gori}, \bibinfo{author}{P.~Gentine},
\newblock \bibinfo{title}{A generative super-resolution model for enhancing tropical cyclone wind field intensity and resolution},
\newblock \bibinfo{journal}{Journal of Geophysical Research: Machine Learning and Computation} \bibinfo{volume}{1} (\bibinfo{year}{2024}) \bibinfo{pages}{e2024JH000375}.
\bibitem[{Oommen et~al.(2024)Oommen, Bora, Zhang, and Karniadakis}]{oommen2024integrating}
\bibinfo{author}{V.~Oommen}, \bibinfo{author}{A.~Bora}, \bibinfo{author}{Z.~Zhang}, \bibinfo{author}{G.~E. Karniadakis},
\newblock \bibinfo{title}{Integrating neural operators with diffusion models improves spectral representation in turbulence modeling},
\newblock \bibinfo{journal}{arXiv preprint arXiv:2409.08477}  (\bibinfo{year}{2024}).
\bibitem[{Hassan et~al.(2023)Hassan, Feeney, Dhruv, Kim, Suh, Ryu, Won, and Chandramowlishwaran}]{hassan2023bubbleml}
\bibinfo{author}{S.~M.~S. Hassan}, \bibinfo{author}{A.~Feeney}, \bibinfo{author}{A.~Dhruv}, \bibinfo{author}{J.~Kim}, \bibinfo{author}{Y.~Suh}, \bibinfo{author}{J.~Ryu}, \bibinfo{author}{Y.~Won}, \bibinfo{author}{A.~Chandramowlishwaran},
\newblock \bibinfo{title}{Bubbleml: a multi-physics dataset and benchmarks for machine learning},
\newblock \bibinfo{journal}{arXiv preprint arXiv:2307.14623}  (\bibinfo{year}{2023}).
\bibitem[{Diakogiannis et~al.(2020)Diakogiannis, Waldner, Caccetta, and Wu}]{diakogiannis2020resunet}
\bibinfo{author}{F.~I. Diakogiannis}, \bibinfo{author}{F.~Waldner}, \bibinfo{author}{P.~Caccetta}, \bibinfo{author}{C.~Wu},
\newblock \bibinfo{title}{Resunet-a: A deep learning framework for semantic segmentation of remotely sensed data},
\newblock \bibinfo{journal}{ISPRS Journal of Photogrammetry and Remote Sensing} \bibinfo{volume}{162} (\bibinfo{year}{2020}) \bibinfo{pages}{94--114}.
\bibitem[{Li et~al.(2018)Li, Xu, Taylor, Studer, and Goldstein}]{li2018visualizing}
\bibinfo{author}{H.~Li}, \bibinfo{author}{Z.~Xu}, \bibinfo{author}{G.~Taylor}, \bibinfo{author}{C.~Studer}, \bibinfo{author}{T.~Goldstein},
\newblock \bibinfo{title}{Visualizing the loss landscape of neural nets},
\newblock \bibinfo{journal}{Advances in neural information processing systems} \bibinfo{volume}{31} (\bibinfo{year}{2018}).
\bibitem[{Chen et~al.(2024)Chen, Liang, Huang, Real, Wang, Pham, Dong, Luong, Hsieh, Lu et~al.}]{chen2024symbolic}
\bibinfo{author}{X.~Chen}, \bibinfo{author}{C.~Liang}, \bibinfo{author}{D.~Huang}, \bibinfo{author}{E.~Real}, \bibinfo{author}{K.~Wang}, \bibinfo{author}{H.~Pham}, \bibinfo{author}{X.~Dong}, \bibinfo{author}{T.~Luong}, \bibinfo{author}{C.-J. Hsieh}, \bibinfo{author}{Y.~Lu}, et~al.,
\newblock \bibinfo{title}{Symbolic discovery of optimization algorithms},
\newblock \bibinfo{journal}{Advances in neural information processing systems} \bibinfo{volume}{36} (\bibinfo{year}{2024}).
\bibitem[{Kingma(2014)}]{kingma2014adam}
\bibinfo{author}{D.~P. Kingma},
\newblock \bibinfo{title}{Adam: A method for stochastic optimization},
\newblock \bibinfo{journal}{arXiv preprint arXiv:1412.6980}  (\bibinfo{year}{2014}).
\bibitem[{Dubey et~al.(2022)Dubey, Weide, O’Neal, Dhruv, Couch, Harris, Klosterman, Jain, Rudi, Messer et~al.}]{dubey2022flash}
\bibinfo{author}{A.~Dubey}, \bibinfo{author}{K.~Weide}, \bibinfo{author}{J.~O’Neal}, \bibinfo{author}{A.~Dhruv}, \bibinfo{author}{S.~Couch}, \bibinfo{author}{J.~A. Harris}, \bibinfo{author}{T.~Klosterman}, \bibinfo{author}{R.~Jain}, \bibinfo{author}{J.~Rudi}, \bibinfo{author}{B.~Messer}, et~al.,
\newblock \bibinfo{title}{Flash-x: A multiphysics simulation software instrument},
\newblock \bibinfo{journal}{SoftwareX} \bibinfo{volume}{19} (\bibinfo{year}{2022}) \bibinfo{pages}{101168}.
\bibitem[{Wei and Zhang(2023)}]{wei2023super}
\bibinfo{author}{M.~Wei}, \bibinfo{author}{X.~Zhang},
\newblock \bibinfo{title}{Super-resolution neural operator},
\newblock in: \bibinfo{booktitle}{Proceedings of the IEEE/CVF Conference on Computer Vision and Pattern Recognition}, \bibinfo{year}{2023}, pp. \bibinfo{pages}{18247--18256}.
\bibitem[{Wang et~al.(2020)Wang, Kashinath, Mustafa, Albert, and Yu}]{wang2020towards}
\bibinfo{author}{R.~Wang}, \bibinfo{author}{K.~Kashinath}, \bibinfo{author}{M.~Mustafa}, \bibinfo{author}{A.~Albert}, \bibinfo{author}{R.~Yu},
\newblock \bibinfo{title}{Towards physics-informed deep learning for turbulent flow prediction},
\newblock in: \bibinfo{booktitle}{Proceedings of the 26th ACM SIGKDD international conference on knowledge discovery \& data mining}, \bibinfo{year}{2020}, pp. \bibinfo{pages}{1457--1466}.
\bibitem[{Chakrabarty and Maji(2019)}]{chakrabarty2019spectral}
\bibinfo{author}{P.~Chakrabarty}, \bibinfo{author}{S.~Maji},
\newblock \bibinfo{title}{The spectral bias of the deep image prior},
\newblock \bibinfo{journal}{arXiv preprint arXiv:1912.08905}  (\bibinfo{year}{2019}).
\bibitem[{Saxe et~al.(2011)Saxe, Koh, Chen, Bhand, Suresh, and Ng}]{saxe2011random}
\bibinfo{author}{A.~M. Saxe}, \bibinfo{author}{P.~W. Koh}, \bibinfo{author}{Z.~Chen}, \bibinfo{author}{M.~Bhand}, \bibinfo{author}{B.~Suresh}, \bibinfo{author}{A.~Y. Ng},
\newblock \bibinfo{title}{On random weights and unsupervised feature learning.},
\newblock in: \bibinfo{booktitle}{Icml}, volume~\bibinfo{volume}{2}, \bibinfo{year}{2011}, p.~\bibinfo{pages}{6}.
\bibitem[{Wang et~al.(2022)Wang, Zheng, Chen, and Wang}]{wang2022anti}
\bibinfo{author}{P.~Wang}, \bibinfo{author}{W.~Zheng}, \bibinfo{author}{T.~Chen}, \bibinfo{author}{Z.~Wang},
\newblock \bibinfo{title}{Anti-oversmoothing in deep vision transformers via the fourier domain analysis: From theory to practice},
\newblock \bibinfo{journal}{arXiv preprint arXiv:2203.05962}  (\bibinfo{year}{2022}).
\bibitem[{Ho et~al.(2020)Ho, Jain, and Abbeel}]{ho2020denoising}
\bibinfo{author}{J.~Ho}, \bibinfo{author}{A.~Jain}, \bibinfo{author}{P.~Abbeel},
\newblock \bibinfo{title}{Denoising diffusion probabilistic models},
\newblock \bibinfo{journal}{Advances in neural information processing systems} \bibinfo{volume}{33} (\bibinfo{year}{2020}) \bibinfo{pages}{6840--6851}.
\bibitem[{Song et~al.(2020)Song, Sohl-Dickstein, Kingma, Kumar, Ermon, and Poole}]{song2020score}
\bibinfo{author}{Y.~Song}, \bibinfo{author}{J.~Sohl-Dickstein}, \bibinfo{author}{D.~P. Kingma}, \bibinfo{author}{A.~Kumar}, \bibinfo{author}{S.~Ermon}, \bibinfo{author}{B.~Poole},
\newblock \bibinfo{title}{Score-based generative modeling through stochastic differential equations},
\newblock \bibinfo{journal}{arXiv preprint arXiv:2011.13456}  (\bibinfo{year}{2020}).
\bibitem[{Song and Ermon(2019)}]{song2019generative}
\bibinfo{author}{Y.~Song}, \bibinfo{author}{S.~Ermon},
\newblock \bibinfo{title}{Generative modeling by estimating gradients of the data distribution},
\newblock \bibinfo{journal}{Advances in neural information processing systems} \bibinfo{volume}{32} (\bibinfo{year}{2019}).
\bibitem[{Karras et~al.(2022)Karras, Aittala, Aila, and Laine}]{karras2022elucidating}
\bibinfo{author}{T.~Karras}, \bibinfo{author}{M.~Aittala}, \bibinfo{author}{T.~Aila}, \bibinfo{author}{S.~Laine},
\newblock \bibinfo{title}{Elucidating the design space of diffusion-based generative models},
\newblock \bibinfo{journal}{Advances in neural information processing systems} \bibinfo{volume}{35} (\bibinfo{year}{2022}) \bibinfo{pages}{26565--26577}.
\bibitem[{Oommen et~al.(2025)Oommen, Bora, Zhang, and Karniadakis}]{Kolmogorov2025Data}
\bibinfo{author}{V.~Oommen}, \bibinfo{author}{A.~Bora}, \bibinfo{author}{Z.~Zhang}, \bibinfo{author}{G.~E. Karniadakis}, \bibinfo{title}{Data for ``integrating neural operators with diffusion models improves spectral representation in turbulence modeling" (kolmogorov flow case)}, \bibinfo{year}{2025}. \URLprefix \url{https://doi.org/10.6084/m9.figshare.28250960.v1}. \DOIprefix\doi{10.6084/m9.figshare.28250960.v1}.
\bibitem[{Settles and Liberzon(2022)}]{settles2022schlieren}
\bibinfo{author}{G.~S. Settles}, \bibinfo{author}{A.~Liberzon},
\newblock \bibinfo{title}{Schlieren and bos velocimetry of a round turbulent helium jet in air},
\newblock \bibinfo{journal}{Optics and Lasers in Engineering} \bibinfo{volume}{156} (\bibinfo{year}{2022}) \bibinfo{pages}{107104}.

\end{thebibliography}

\appendix
\section{Training strategies and ResUNet prediction results}
\label{appA}
All the models were trained for \(\sim\)1000 epochs (convergence typically happened earlier). The initial learning rate was set to \(8\times 10^{-4}\) and it was reduced after the first 700 epochs using a linear step scheduler. We used GELU activation function and group normalization after convolutional layers. Lion optimizer with weight decay of 0.02 to 0.1 were used, depending on the neural operator size. Batch size of 4 or 8 was used, depending on the neural operator size. We found that gradient clipping at maximum gradient norm of 0.4 to 1 (depending on neural operator size) helps with the optimization. Our preliminary findings showed better results with the Lion optimizer compared to Adam and AdamW optimizers. Therefore, all the trainings for this work were conducted with the Lion optimzier. For all the neural operators, the number of layers in the encoder and decoder were kept constant and the number of parameters at each layer was modified to change the neural operator size. 

\begin{figure}[H]
    \centering
    \includegraphics[width=1\textwidth]{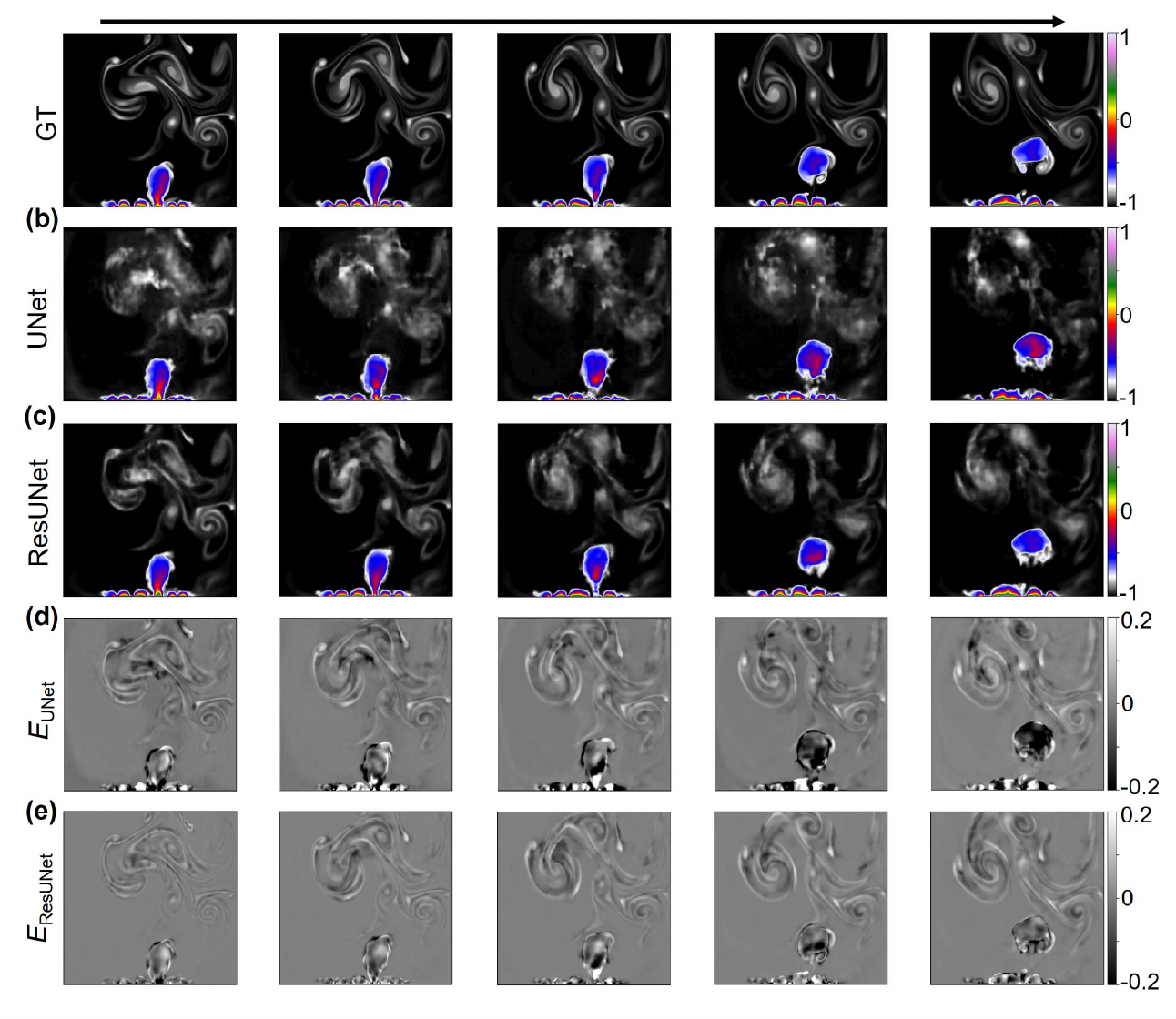}
    \caption{\textbf{Example of subcooled pool boiling temperature prediction results by neural operators.} (a) Ground truth results, (b) UNet prediction results, (c) ResUNet prediction results, (d) UNet prediction errors, (e) ResUNet prediction errors. The results show five time-step predictions from left to right.}
    \label{Figure A12}
\end{figure}

\section{Boundary RMSE, Bubble RMSE, and Spectral \\Errors}
\label{appB}
Boundary RMSE (BRMSE) for a single sample and time-step is defined by calculating the errors only at the boundaries of the domain:
\begin{equation}
    \text{BRMSE} = \sqrt{\frac{1}{|\partial \Omega|} \sum_{\mathbf{x}_i \in \partial \Omega} \left( \hat{T}_i - T_i \right)^2},
\end{equation}
where \(\mathbf{x}_i \in \partial \Omega\) specifies the points at the boundaries, \(\hat{T}_i\) is the predicted temperature, and \(T_i\) is the actual temperature. Similarly, bubble RMSE is defined by calculating the errors only within the bubble areas. These areas are specified through a level-set function in the simulations.
\begin{equation}
    \text{Bubble RMSE} = \sqrt{\frac{1}{|\Omega_{\text{bubble}} \cup \partial \Omega_{\text{bubble}}|} \sum_{\mathbf{x}_i \in \Omega_{\text{bubble}} \cup \ \partial \Omega_{\text{bubble}}} \left( \hat{y}_i - y_i \right)^2},
\end{equation}
where \(\mathbf{x}_i \in \Omega_{\text{bubble}}\) and \(\partial \Omega_{\text{bubble}}\) specify the points inside the bubble areas and at the interfaces, respectively.\\

The spectral errors in each of the low, mid, and high-frequency bands are defined as follows:
\begin{equation}
    F_{\text{band}} = \sqrt{\frac{1}{N_{\text{band}}} \sum_{k \in \text{band}} \left| \mathcal{F}(T)(\mathbf{k}) - \mathcal{F}(\hat{T})(\mathbf{k}) \right|^2} \quad \text{,} \quad \text{band} \in \{ \text{low}, \text{mid}, \text{high} \},
\end{equation}

where \textit{k} is the spatial frequency component of the Fourier transformed solutions, \(\mathcal{F}\) denotes the Fourier transform, and \({N_{\text{band}}}\) specifies the number of components at each frequency band. The low, mid, and high bands may be defined differently based on the underlying dataset and the amount of high-frequency components. In this work, these bands were set to the first 2\%, the first 6.2\% excluding the low band components, and the last 93.8\% of the components.\\

Similarly, the energy spectrum error, showing the energy spectra misalignment at each frequency band is defined as follows:
\begin{equation}
\mathcal{E}_{F_{\text{band}}} = \sqrt{\frac{1}{N_{\text{band}}} \sum_{k \in \text{band}} \left( \Big| \mathcal{F}(T)(\mathbf{k}) \Big|^2 - \Big| \mathcal{F}(\hat{T})(\mathbf{k}) \Big|^2 \right)^2 } \quad \text{,} \quad \text{band} \in \{ \text{low}, \text{mid}, \text{high} \},
\end{equation}

where \(\mathcal{E}\) denotes the energy spectrum error.

\section{Summary of subcooled pool boiling prediction results with HFS-enhanced NO}
\label{appC}
In this work, we tested different variants of ResUNet by varying number of parameters in the range of \(\sim\)2 millions to \(\sim\)16 millions. In the following table, we summarized the results of the two of the models (smallest and largest models), trained with optimal hyperparameters. Note that the same hyperparameters were used for training a neural operator with and without HFS. The parameters were first optimized for the NO without HFS and the same set of parameters were used for training the HFS-enhanced NO. The results of the other models are not included in this table for easier comparison and interpretation. We refer the reader to Figure 4 for observing the effect of HFS on all the tested models. Similar to the rest of the paper, the results are based on five time-step predictions.

\begin{table}[H]
    \caption{\textbf{Subcooled pool boiling temperature prediction errors with neural operator (NO) with and without high-frequency scaling (HFS)} The columns correspond to the metrics, NO with \(\sim\)1.7 millions parameters,  HFS-enhanced NO with \(\sim\)1.7 millions parameters, NO with \(\sim\)16.2 millions parameters, and HFS-enhanced NO with \(\sim\)16.2 millions parameters.}
    \label{Table B}
    \centering
    \resizebox{\textwidth}{!}{%
    \begin{tabular}{c|c|c|c|c}
    \hline
        & \textbf{NO, 1.7 M}
        & \textbf{NO+HFS, 1.7 M} & \textbf{NO, 16.2 M}
        & \textbf{NO+HFS, 16.2 M}\\
    \hline
    \textbf{Rel. Error} &0.0414 &0.0333 &0.0251 &0.0238 \\ \hline
    \textbf{RMSE} &0.0403 &0.0324 &0.0244 &0.0232 \\ \hline
    \textbf{BRMSE} &0.0973 &0.0729 &0.0562 &0.0505 \\ \hline
    \textbf{Bubble RMSE} &0.1924 &0.143 &0.109 &0.0985 \\ \hline
    \textbf{Max\textsubscript{mean}} &1.019 &0.897 &0.685 &0.656 \\ \hline
    \textbf{\textit{F}\textsubscript{low}} &0.323 &0.237 &0.212 &0.141 \\ \hline
    \textbf{\textit{F}\textsubscript{mid}} &0.282 &0.218 &0.185 &0.148 \\ \hline
    \textbf{\textit{F}\textsubscript{high}} &0.0476 &0.0400 &0.0392 &0.0296 \\ \hline
    \textbf{Parameters [Millions]} &1.711& 1.712 &16.263 &16.268\\ \hline
    \end{tabular}%
    }
\end{table}

\section{Saturated pool boiling prediction results}
\label{appD}
Saturated pool boiling dataset involves less complexity due to lower high-frequency components and small scale features. Therefore, a well-optimized NO without HFS can successfully resolve the solutions. However, HFS still enhances the prediction accuracies, especially at bubble areas. The following figure demonstrates an example of predictions using NO and HFS-enhanced NO for saturated pool boiling dataset. Generally, the errors are much smaller than subcooled pool boiling predictions. However, it can be seen that the errors in the regions with departed bubbles are reduced with HFS-enhanced NO.
\begin{figure}[H]
    \centering
    \includegraphics[width=1\textwidth]{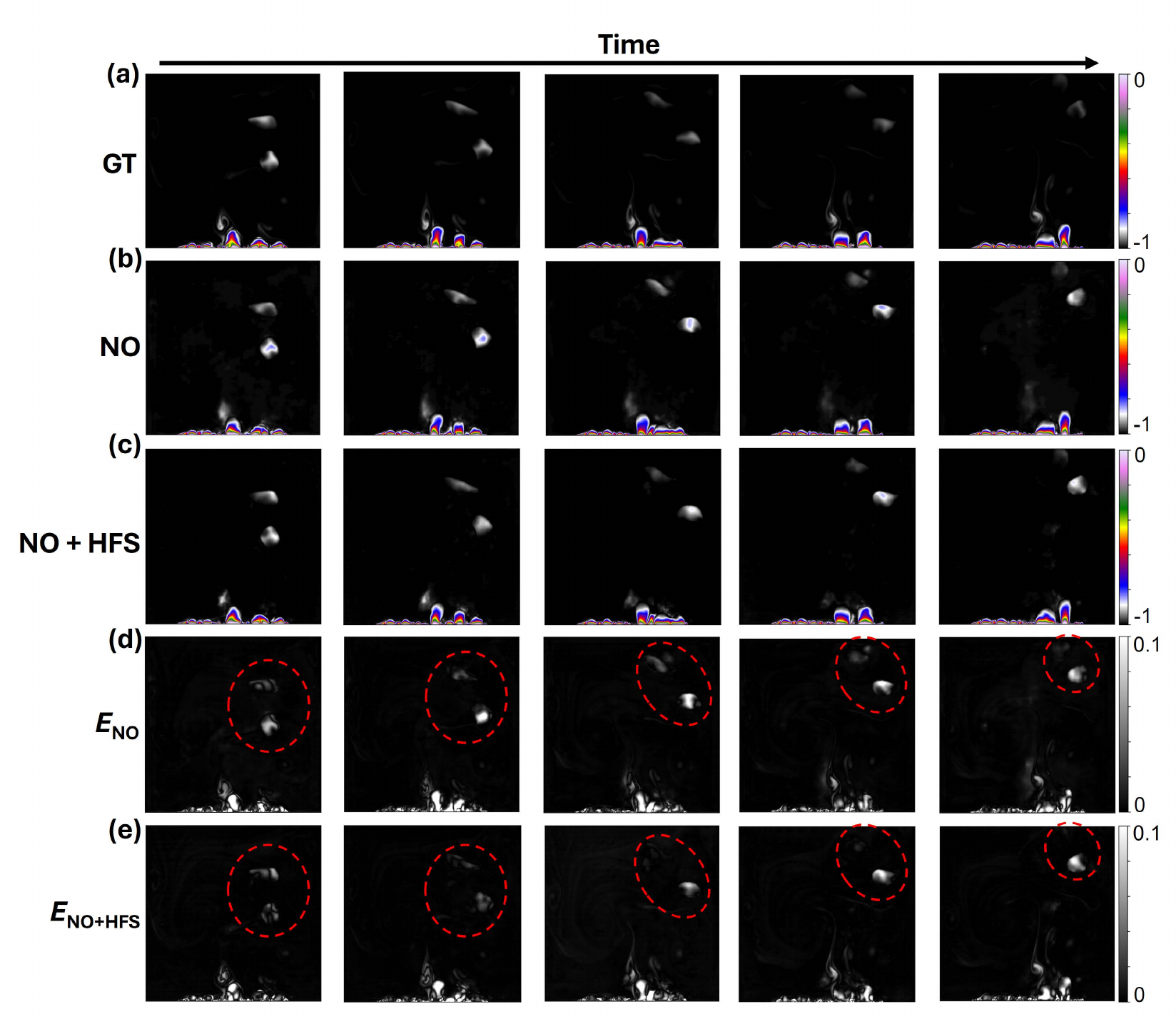}
    \caption{\textbf{Examples of saturated pool boiling temperature prediction results by NO and HFS-enhanced NO} (a) Ground truth (GT) results. (b) NO predictions. (c) NO + HFS predictions. (d) Absolute prediction errors of NO ($E_{\text{NO}}$). (e) Absolute prediction errors of NO + HFS ($E_{\text{NO+HFS}}$). The results are shown for five time-step predictions from left to right. The departed bubbles areas are circled (dashed red circles) in error maps for easier interpretation and comparison. The results are based on a NO with \(\sim\) 3.5 millions parameter.}
    \label{Figure C12}
\end{figure}

To further investigate if HFS can enhance the predictions with smaller NO on this simpler dataset, we trained another NO with the same structure (ResUNet) but with only \(\sim\) 0.6 millions parameters with and without HFS. Consistent with previous results, HFS enhanced the predictions by reducing the field errors such as RMSE and bubble RMSE as well the spectral errors. The prediction results of saturated pool boiling dataset using two different NOs with and without HFS are summarized in the following table. Similar to the rest of the paper, the results are based on five time-step predictions.

\begin{table}[H]
    \caption{\textbf{Saturated pool boiling temperature prediction errors of NO with and without HFS.} The columns correspond to the metrics, NO with \(\sim\)0.6 millions parameters, HFS-enhanced NO with \(\sim\)0.6 millions parameters, NO with \(\sim\)3.5 millions parameters, and HFS-enhanced NO with \(\sim\)3.5 millions parameters.}
    \label{Table B}
    \centering
    \resizebox{\textwidth}{!}{%
    \begin{tabular}{c|c|c|c|c}
    \hline
        & \textbf{NO, 0.6 M}
        & \textbf{NO+HFS, 0.6 M} & \textbf{NO, 3.5 M}
        & \textbf{NO+HFS, 3.5 M}\\
    \hline
    \textbf{Rel. Error} &0.0173 &0.0165 &0.0149 &0.0145 \\ \hline
    \textbf{RMSE} &0.0171 &0.0164 &0.0148 &0.0144 \\ \hline
    \textbf{BRMSE} &0.0462 &0.0450 &0.0364 &0.0355\\ \hline
    \textbf{Bubble RMSE} &0.0918 &0.0898 &0.0726 &0.0692\\ \hline 
    \textbf{Max\textsubscript{mean}} &0.592 &0.595 &0.553 &0.544\\ \hline
    \textbf{\textit{F}\textsubscript{low}} &0.0964 &0.0835 &0.0745 &0.0736\\ \hline
    \textbf{\textit{F}\textsubscript{mid}} &0.1086 &0.0998 &0.0919 &0.0855\\ \hline
    \textbf{\textit{F}\textsubscript{high}} &0.0209 &0.0208 &0.0182 &0.0180\\ \hline
    \textbf{Parameters [Millions]} &0.614& 0.615 &3.480 &3.481\\ \hline
    \end{tabular}%
    }
    
\end{table}

\section{Visualization of latent space}
\label{appE}
\begin{figure}[H]
    \centering
    \includegraphics[width=1\textwidth]{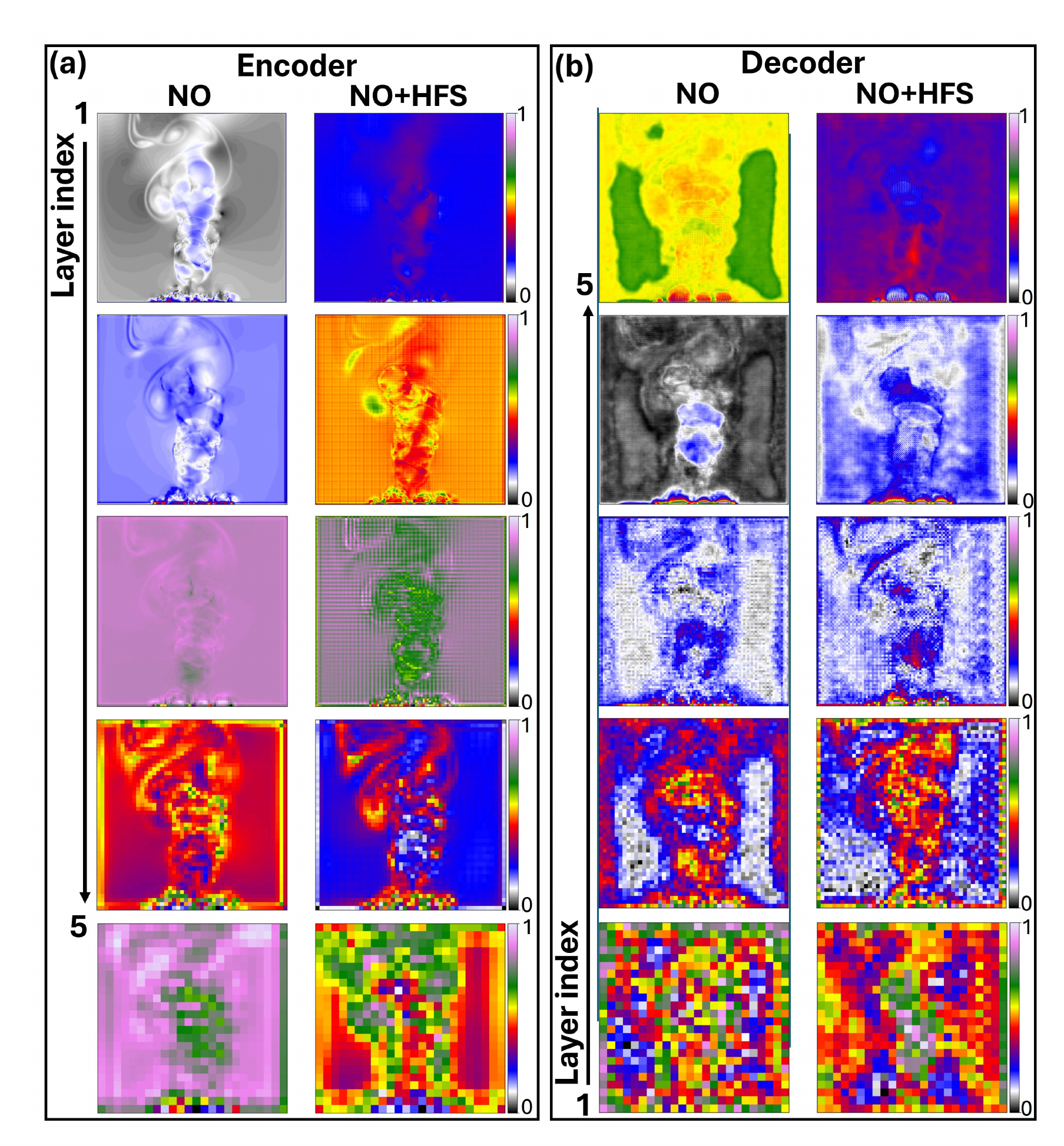}
    \caption{\textbf{Effect of HFS on the latent space mean features.} (a) Mean latent feature maps in decoder (downsampling) with five layers. (b) Mean latent feature maps in decoder (upsampling) with five layers. The results are based on a NO with \(\sim\) 16 millions parameters.}
    \label{Figure D13}
\end{figure}

\section{Additional visualizations of the subcooled pool boiling predictions}
\label{appF}
\begin{figure}[H]
    \centering
    \includegraphics[width=1\textwidth]{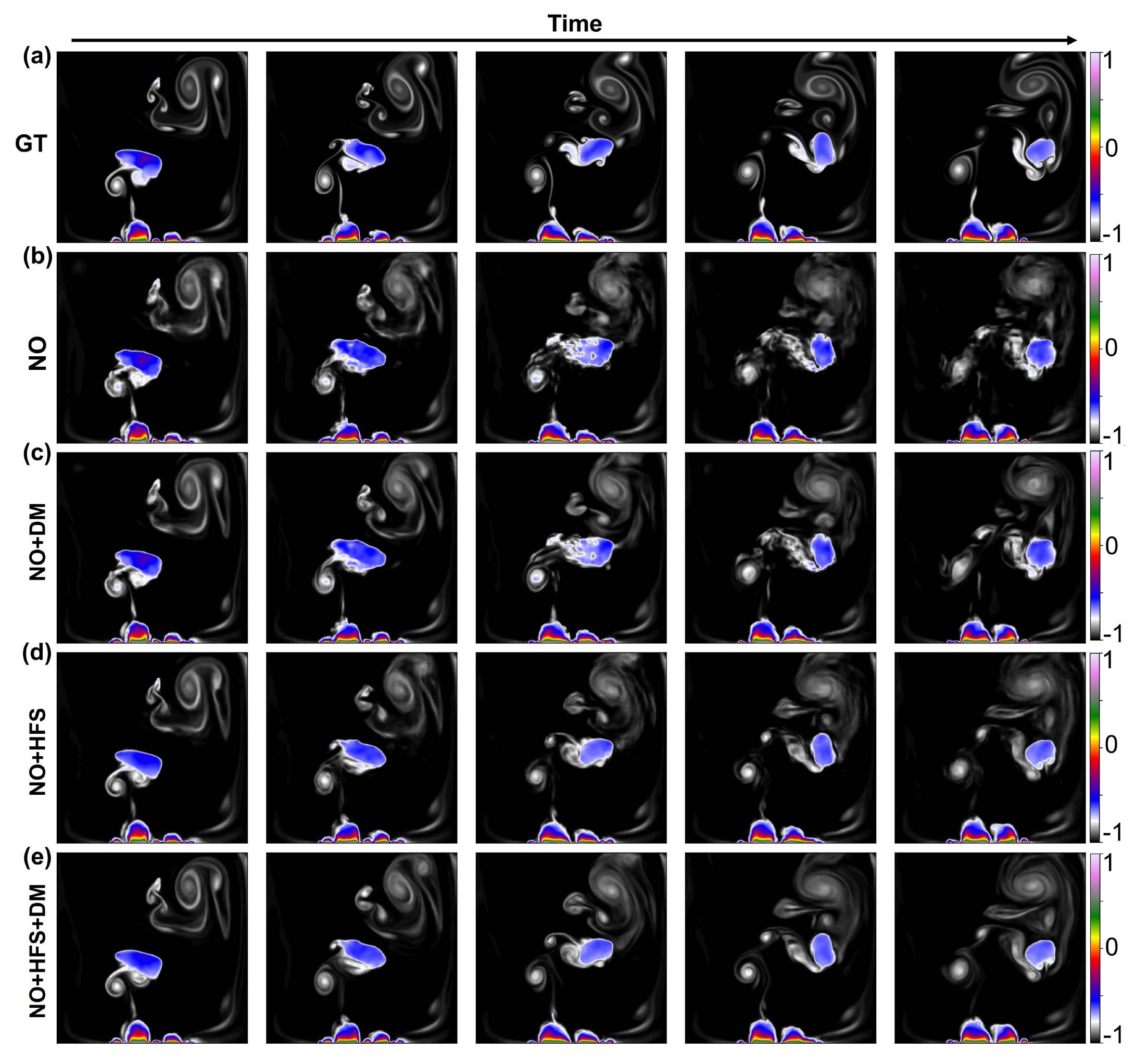}
    \caption{\textbf{Examples of subcooled pool boiling prediction results by DM integrated with NO and HFS-enhanced NO.} (a) Ground truth (GT) results. (b) NO predictions. (c) NO + DM predictions. (d) NO + HFS predictions. (e) NO + HFS + DM predictions. The results are shown for five time-step predictions from left to right.}
    \label{Figure D1}
\end{figure}

\section{Optimized scaling parameters, \(\lambda_{DC}\) and \(\lambda_{HFC}\)}
\label{appG}
The following figure demonstrates the learned \(\lambda_{DC}\) and \(\lambda_{HFC}\) across all the feature maps in the latent space of the encoder and decoder. The results are based on the training of a HFS-enhanced NO with \(\sim\) 1.7 million parameters for the subcooled pool boiling problem.

\begin{figure}[H]
    \centering
    \includegraphics[width=0.9\textwidth]{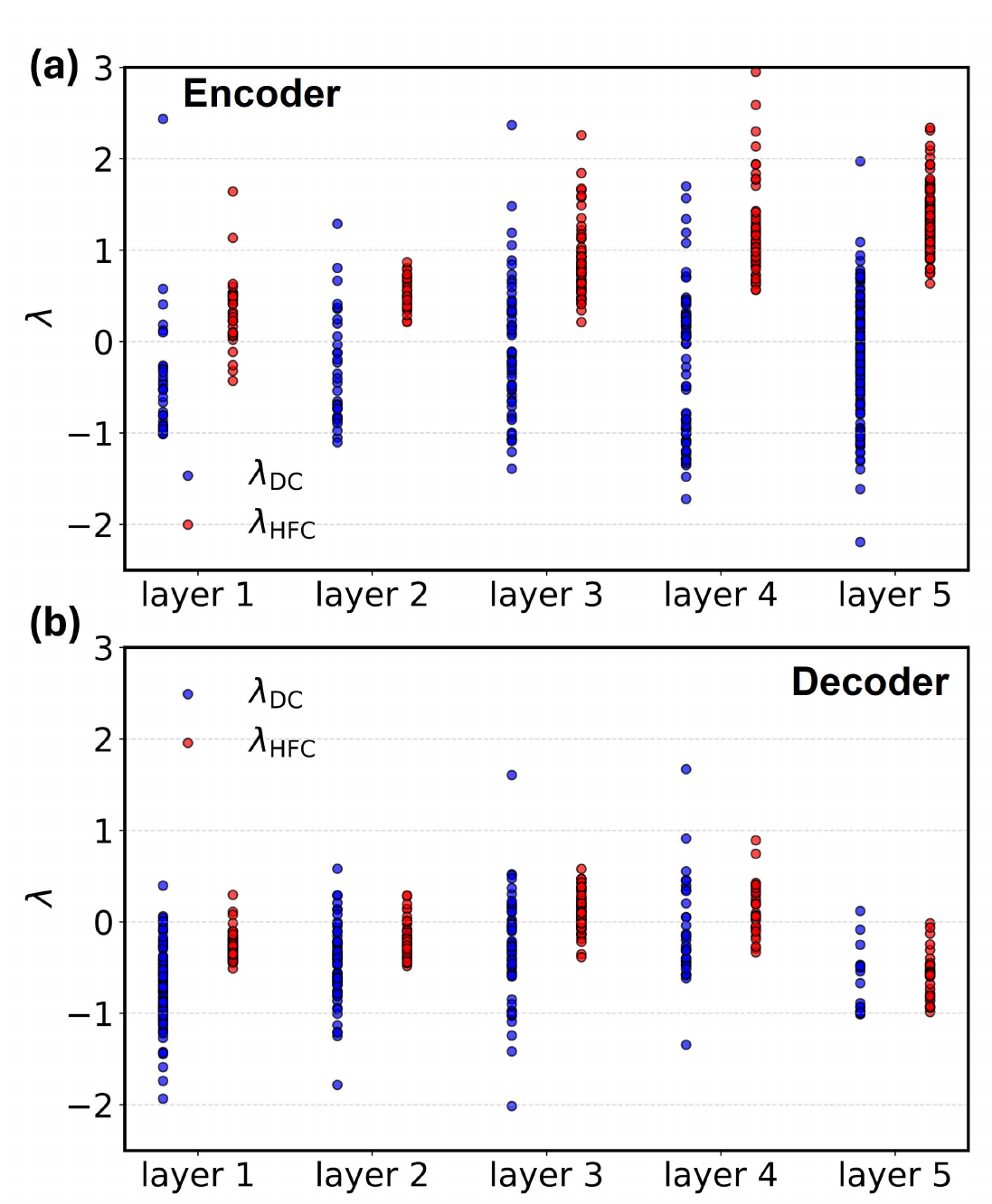}
    \caption{(a) learned values of \(\lambda_{DC}\) and \(\lambda_{HFC}\) in the encoder of the NO. (b) learned values of \(\lambda_{DC}\) and \(\lambda_{HFC}\) in the decoder of NO. Layers start from highest spatial resolution to the lowest in the encoder and vice versa for the decoder.}
    \label{Figure G15}
\end{figure}

\section{Kolmogorov flow prediction results}
\label{appH}
The vorticity formulation of the unsteady 2D incompressible Navier-Stokes equation for a viscous and incompressible fluid with the Kolmogorov forcing term is given as follows, where  \(\omega\) is the vorticity, \(u\) is the velocity vector, and \(\nu\) is the kinematic viscosity.

\begin{equation}
\left\{
\begin{array}{ll}
\partial_t \omega + \mathbf{u} \cdot \nabla \omega = \nu \Delta \omega + f(x, y),  
& (x, y) \in (0, 2\pi)^2, \ t \in (0, t_{\text{final}}] \\ 
f(x, y) = \chi (\sin(2\pi(x + y)) + \cos(2\pi(x + y))),  
& (x, y) \in (0, 2\pi)^2 \\ 
\nabla \cdot \mathbf{u} = 0,  
& (x, y) \in (0, 2\pi)^2, \ t \in (0, t_{\text{final}}] \\ 
\omega(x, y, 0) = \omega_0,  
& (x, y) \in (0, 2\pi)^2
\end{array}
\right.
\end{equation}

In this study, we used \(\chi\) = 0.1, \(\nu = 10^{-5}\), and periodic boundary conditions. The vorticity initial condition was sampled from a Gaussian random field according to the distribution \(\mathcal{N}(0,14^{0.5}(-\Delta + 196I)^{-1.5})\). The following figure demonstrate an example of the prediction results of the neural operator with and without HFS.

\begin{figure}[H]
    \centering
    \includegraphics[width=1\textwidth]{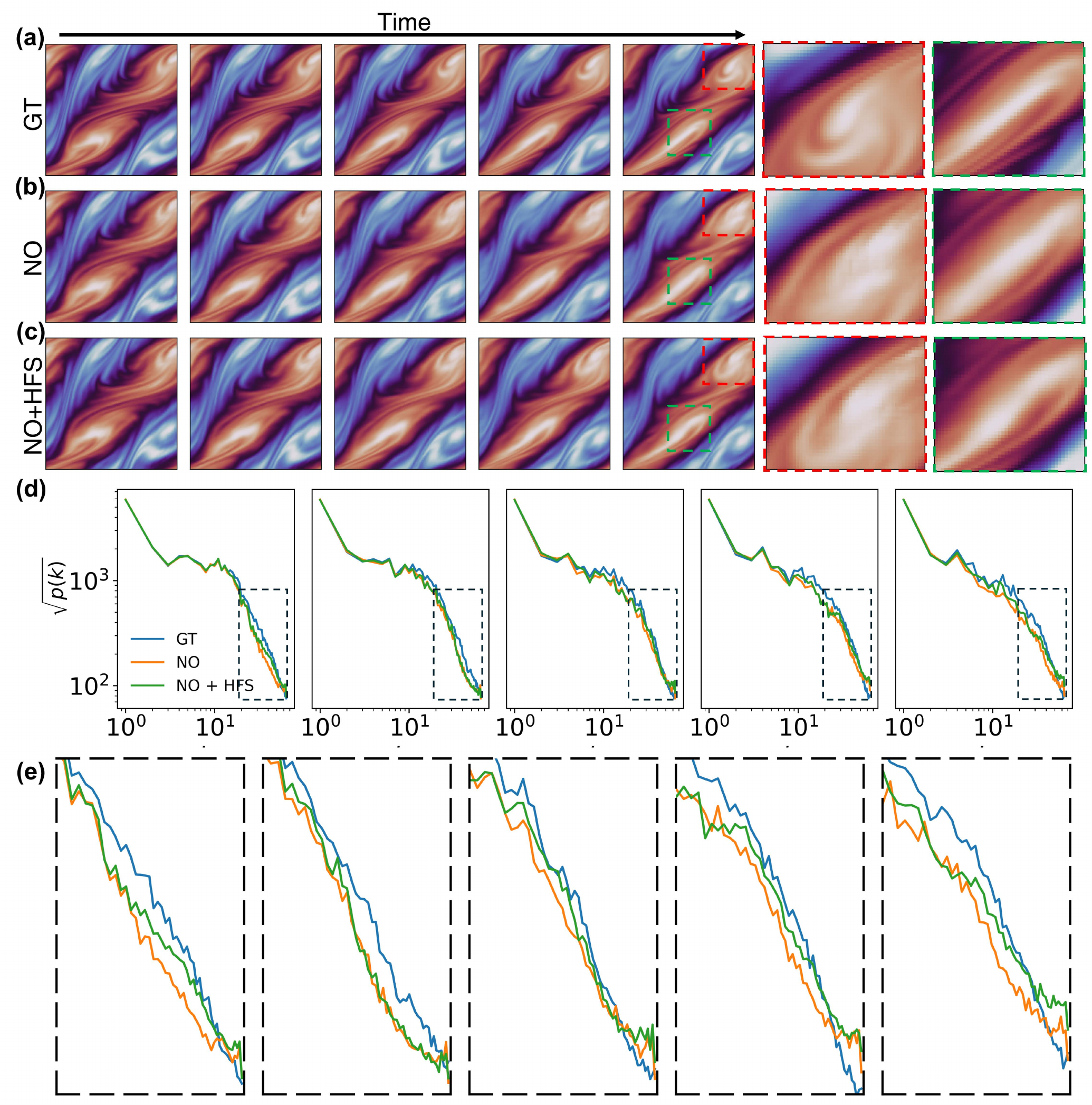}
    \caption{\textbf{2D Kolmogorov flow prediction results.} (a) Ground truth solutions. (b) NO predictions. (c) HFS-enhanced NO predictions. (d) The corresponding energy spectra (\((p(k))\) for predictions at each time-step. (e) Zoomed-in view of energy spectra showing only the high wavenumbers for better visualization of the differences. The legends in (d) are applicable to (e) as well.}
\end{figure}

\end{document}